\documentclass[twoside,11pt]{article}

\usepackage{jmlr2e, rawfonts}
\usepackage{natbib}

\usepackage[utf8]{inputenc} % allow utf-8 input
\usepackage[T1]{fontenc}    % use 8-bit T1 fonts
\usepackage{url}            % simple URL typesetting
\usepackage{booktabs}       % professional-quality tables
\usepackage{amsfonts}       % blackboard math symbols
\usepackage{nicefrac}       % compact symbols for 1/2, etc.
\usepackage{microtype}      % microtypography

\usepackage{graphicx} % more modern
\usepackage{subcaption}
\usepackage[rightcaption]{sidecap}
\usepackage{tabularx}
\usepackage{pbox}
\usepackage{csquotes}
\usepackage{booktabs}
\usepackage{xcolor}
\usepackage{wrapfig}
\usepackage{caption}
\sidecaptionvpos{figure}{c}
\usepackage[framemethod=TikZ]{mdframed}
\usepackage{floatrow}

\usepackage{amsmath}
\usepackage{mathtools}
\usepackage{thmtools,thm-restate}

\usepackage{algorithm}
\usepackage{algorithm,algorithmicx}
\usepackage[noend]{algpseudocode}

\usepackage{enumitem}
\usepackage{subfiles}

\graphicspath{{./figures/}{../figures}}

\usepackage[most]{tcolorbox}

\newtcbtheorem{Remark}{\bfseries Remark}{enhanced,drop shadow={black!50!white},
  coltitle=white,
  top=0.1in,
}{summary}

\def\Nat{{\rm I\kern\pIR N}}

\newcommand{\EE}[1]{\exptE\left[#1\right]}

\newcommand{\defeq}{\overset{\text{\tiny def}}{=}}

\newcommand{\PerfRV}{M}
\newcommand{\numest}{n}
\newcommand{\numrepeats}{m}

\def\vec0{{\boldsymbol{0}}}

\newcommand{\beq}{\begin{equation}}
\newcommand{\eeq}{\end{equation}}
\newcommand{\beqa}{\begin{eqnarray}}
\newcommand{\eeqa}{\end{eqnarray}}
\newcommand{\beqan}{\begin{eqnarray*}}
\newcommand{\eeqan}{\end{eqnarray*}}
\newcommand{\ben}{\begin{eqnarray*}}
\newcommand{\een}{\end{eqnarray*}}

\renewcommand{\EE}[2]{\mathbb{E}_{#1\!\!}[#2]}

\newcommand{\CEE}[3]{\EE{#1}{{#2}~\middle\vert~{#3}}}
\renewcommand{\CEE}[3]{\EE{#1}{{#2}\mid{#3}}}
\def\CE#1#2{\CEE{\,}{#1}{#2}}

\usepackage[most]{tcolorbox}
\newcounter{myexample}
\usepackage{xparse}
\usepackage{lipsum}

\tikzset{text=white}
\newenvironment{summary}[2][]{%
\ifstrempty{#1}%
{\mdfsetup{%
coltext=White,
frametitle={%
\tikz[baseline=(current bounding box.east),outer sep=0pt]
\node[anchor=east,rectangle,fill=white,colback=white,]
{\strut };}}
}%
{\mdfsetup{%
frametitle={%
\tikz[baseline=(current bounding box.east),outer sep=0pt]
\node[anchor=east,rectangle,fill=blue!75!black]
{\strut~#1};}}%
}%
\mdfsetup{innertopmargin=10pt,linecolor=blue!75!black,%
linewidth=2pt,topline=true,%
frametitleaboveskip=\dimexpr-\ht\strutbox\relax
}
\begin{mdframed}[]\relax%
\label{#2}}{\end{mdframed}}

\def\exampletext{Example}

\newenvironment{myexample}[1][\unskip]
{
\colorlet{colexam}{red!55!black} % Global example color
\newtcolorbox[use counter=example]{examplebox}{%
    empty,% Empty previously set parameters
    title={\exampletext: #1},% use \thetcbcounter to access the example counter text
    attach boxed title to top left,
       minipage boxed title,
    boxed title style={empty,size=minimal,toprule=0pt,top=4pt,left=3mm,overlay={}},
    coltitle=colexam,fonttitle=\bfseries,
    before=\par\medskip\noindent,parbox=false,boxsep=0pt,left=3mm,right=0mm,top=2pt,breakable,pad at break=0mm,
       before upper=\csname @totalleftmargin\endcsname0pt, % Use instead of parbox=true. This ensures parskip is inherited by box.
    overlay unbroken={\draw[colexam,line width=.5pt] ([xshift=-0pt]title.north west) -- ([xshift=-0pt]frame.south west); },
    overlay first={\draw[colexam,line width=.5pt] ([xshift=-0pt]title.north west) -- ([xshift=-0pt]frame.south west); },
    overlay middle={\draw[colexam,line width=.5pt] ([xshift=-0pt]frame.north west) -- ([xshift=-0pt]frame.south west); },
    overlay last={\draw[colexam,line width=.5pt] ([xshift=-0pt]frame.north west) -- ([xshift=-0pt]frame.south west); },%
    }
\begin{examplebox}}
{\end{examplebox}\endlist}

\newcommand{\adam}[1]{\PackageError{comments}{Please Set your color with the colour package.}{extra help}}

\def\biblio{\bibliographystyle{plainnat}\bibliography{thesis}}

\usepackage{tcolorbox}

\usepackage{lastpage}
\jmlrheading{25}{2024}{1-\pageref{LastPage}}{2/23; Revised10/24}{10/24}{23-0183}{Patterson, Neumann, White, White}
\ShortHeadings{Empirical Design in Reinforcement Learning}{Patterson, Neumann, White, White}
\firstpageno{1}

\begin{document}
\def\biblio{}

\title{Empirical Design in Reinforcement Learning}

\author{\name Andrew Patterson \email ap3@ualberta.ca \\
	\name Samuel Neumann  \email sfneuman@ualberta.ca \\
	\name Martha White$^\dagger$ \email whitem@ualberta.ca \\
	\name Adam White$^\dagger$  \email amw8@ualberta.ca \\
       \addr
      Department of Computing Science and Alberta Machine Intelligence Institute (Amii) \\
       University of Alberta, Edmonton, Canada\\
           $^\dagger$Canada CIFAR AI Chair
}

\editor{George Konidaris}

\maketitle

\begin{abstract}%
Empirical design in reinforcement learning is no small task.
%There is no justification for conducting a flawed experiment and providing misleading claims about the results.
Running good experiments requires attention to detail and at times significant computational resources.
While compute resources available per dollar have continued to grow rapidly, so have the scale of typical experiments in reinforcement learning. It is now common to benchmark agents with millions of parameters against dozens of tasks, each using the equivalent of 30 days of experience. The scale of these experiments often conflict with the need for statistical evidence, especially when comparing algorithms. Recent studies have highlighted how popular algorithms are sensitive to hyperparameter settings and implementation details, and that common empirical practice leads to weak statistical evidence \citep{machado2018revisiting,henderson2018deep}. %Here we take this one step further.

This manuscript represents both a call to action, and a comprehensive resource for how to do good experiments in reinforcement learning.
%We begin with a case study investigating how poor methodological choices in prior work suggested that XYZ, whereas careful study reveals the opposite. From there
In particular, we cover: the statistical assumptions underlying common performance measures, how to properly characterize performance variation and stability, hypothesis testing, special considerations for comparing multiple agents, baseline and illustrative example construction, and how to deal with hyperparameters and experimenter bias. Throughout we highlight common mistakes found in the literature and the statistical consequences of those in example experiments.
The objective of this document is to provide answers on how we can use our unprecedented compute to do good science in reinforcement learning, as well as stay alert to potential pitfalls in our empirical design.
\end{abstract}


%\newpage
%\tableofcontents
%\newpage

\section{Reinforcement Learning, an Empirical Science}\label{sec:intro}

Running a good experiment in reinforcement learning is difficult.
There are many decisions to be made:
  How long should you run your agent?
  Should you count the number of episodes or number of steps?
  Should performance be measured online or offline with test trials?
  How should you measure and aggregate performance?
  Do we use rules of thumb to set hyperparameters or some systematic search?
  What are the right baseline algorithms to compare against?
  Which environments should you use?
  What does good learning even look like in a given environment?
The answer to each question can greatly impact the credibility and utility of the result, ranging from insightful to down-right misleading.

The task of evaluating a reinforcement learning agent is complicated by the fundamental aspect that makes the problem interesting: an agent interacting with an environment.
Unlike supervised learning, reinforcement learning experiments are online and interactive.
The agent---a program---generates its own training data by interacting with the environment---another program---and the quality of the data depends on what the agent learned previously.
This interaction makes fair comparisons and scientific reproducibility major challenges in reinforcement learning.
Many of the ideas from classical machine learning such as train and test splits, overfitting, cross-validation, and model selection are either different or non-existent in reinforcement learning.
It is not surprising that the community is currently wrestling with the consequences of limited reproducibility, experimenter bias, unreliable algorithms, and exaggerated performance claims.

The field of reinforcement learning is experiencing rapid growth and many of the issues we see today are expected of a growing field.
Historically, the community was much smaller than other branches of machine learning and the scale of most experiments was limited: a large scale experiment may have consisted of a half dozen state dimensions and thousands of episodes.
Before 2014 (with the development of DQN), researchers were likely to begin by replicating experiments from the Sutton and Barto textbook (2018) and then extending and innovating from there.
This was perhaps serendipitous as Sutton and Barto spent decades refining their experimental insights: learning from animal learning experiments and from teaching the textbook year after year.
Most researchers were starting from an excellent empirical foundation.
Historically, large scale experiments in reinforcement learning such as TD-Gammon \citep{tesauro1995temporal} and work in robotics were demonstrations highlighting what was \emph{possible} with reinforcement learning, without attempting to make strong scientific claims.

This document makes a distinction between scientific studies in reinforcement learning and demonstrations of (impressive) engineered systems. Both play an important role in reinforcement learning research, but can be detrimental when conflated. Demonstrations can be seen as exploratory science, probing the edges of what is known or demonstrating the capabilities of existing algorithms. Scientific studies, on the other hand, aim to obtain a deeper understanding of our systems and algorithms; typically by posing clear and falsifiable hypotheses and controlling for confounding effects.
\textbf{The aim should not be to show an algorithm is good, but rather understand an algorithm's properties, potentially relative to other algorithms.}\footnote{Of course, an important part of understanding is also theoretical analysis. Experiments and analysis go hand-in-hand, in that they both aim to provide understanding, using different tools.%Sometimes the right answer is to use analysis, rather than experiments, or to do both. 
This document is focused on good experiments, but does not suggest that this is the only route to understanding our algorithms.}

Given the immense growth of the field, there is an emerging need to more clearly articulate and even develop better empirical practices in reinforcement learning.
There is certainly a greater variety of researchers operating in reinforcement learning currently.
Many come from other fields of machine learning and neuroscience, bringing with them different expectations, practices, and rules of thumb.
Much of the widely bemoaned poor empirical practices \citep{henderson2018deep,jordan2020evaluating,agarwal2021deep,colas2018how} could be due to mistakenly applying practices common in other communities.

In this document we aim to provide a cookbook or how-to guide for running good experiments in reinforcement learning.
We will walk the budding reinforcement learning empiricist through important design decisions, common mistakes, and hidden biases.
We will provide numerical examples of the consequences of bad decisions and illustrate what clear results and fair comparisons look like.
In some cases we convey rules of thumb and sources of bias hard learned over decades of experience in the field.
Naturally, we can never cover all the key decisions and the list of bad practice will be incomplete and ever-growing.
Regardless, our ambition is to provide (1) a reference on how to run good experiments in reinforcement learning for those new to the field, and (2) additional insight and examples so that this cookbook may be useful to the seasoned researcher as well.

We begin by discussing the complexity of a first experiment one might run: evaluating a single agent on a single environment in Section \ref{obs-exps}. The section introduces key concepts about what to measure and how to aggregate performance in Section \ref{sec_first}. Then we get more technical, looking at sources of variability in our experiments (Section \ref{sec_single_more}), and how to make statistically significant claims (Sections \ref{var-of-perf}, \ref{sec_ci}). In Section \ref{sec_hypers} we address how to deal with hyperparameters in our experiments. After that we move to comparing multiple agents, and the additional nuances that arise in Section \ref{sec_comparing}. In Section \ref{sec_environments} we discuss some considerations when selecting environments for your experiments. In Section \ref{sec:case-study}, we attempt to recreate a previous result and demonstrate how to use the strategies explained here to improve on the previous experiment design.
We conclude with a summary of common errors made in experiments in Section \ref{errors}. 

\begin{figure}[htb]
	\centering
	\includegraphics[width=1.0\linewidth]{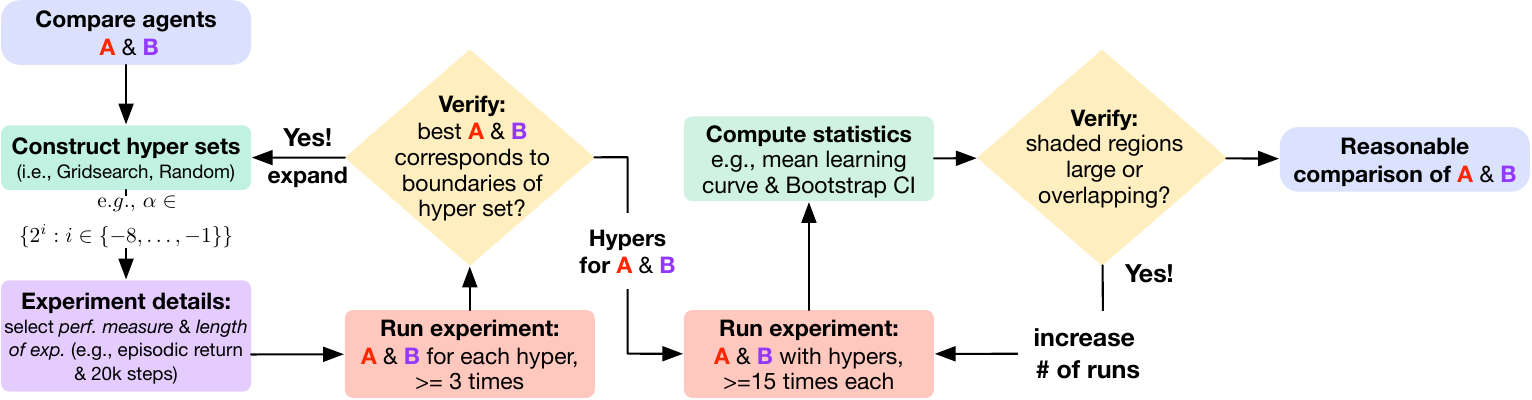}
	\caption{
    \textbf{Two-stage approach to comparing two algorithms.} The goal of this experimental workflow is to progress left to right. Good choices in the colored boxes should limit the experiment rerunning sometimes forced by the yellow decision diamonds. The same basic process is used with automatic hyperparameter optimization algorithms (HOA), except the HOA chooses what hyperparameters to test from a specified range (instead of a set) and sometimes early stopping is used (during the first red {\em Run experiment} box). You should still use multiple runs to evaluate the hyperparameters within the HOA and after it is done you should check if the best hyperparameters are at the edge of the ranges. From there, one would proceed with {\em Run experiment}---second red box---and continue with the workflow.
	  Each stage of the work flow is discussed in detail in the text; we link each for convenience. \textbf{Section \ref{obs-exps}:} discusses how to decide on key experiment details (in \ref{sec_steps}), the basics on how to run an experiment (in \ref{sec_first}), plot learning curves and confidence intervals (in \ref{sec_ci}), and decide if we need more runs (in \ref{sec_moreruns}). \textbf{Section \ref{sec_hypers}:} discusses how to construct hyperparameter sets, and how to determine if you need to expand the hyperparameter set (in \ref{hyper_overall}).
    \textbf{Section \ref{sec_twoalgs}:} discusses ways to compare two (or more) algorithms, including how to detect if the changes you have made to a baseline significantly improve performance.
  }
	\label{fig:2-stage}
\end{figure}

With such a large document, it is often useful to provide a high-level summary.
Figure~\ref{fig:2-stage} provides a flowchart visualizing an experimental procedure with references to associated sections.
Specifically, we visualize a common two-stage procedure which represents sensible empirical workflow and should lead to reasonable conclusions.

This document is educational, but does in fact contain a variety of new results. These results are used to support methodological proposals throughout this work. For easier reference, we list those novel findings in Appendix \ref{app_contributions}. We also include a summary list of common errors and pitfalls at the end of this document, in Section \ref{errors}.
% and a flow chart visualizing a good experimental workflow in Section \ref{flowchart}.

\biblio

\section{Observational experiments}
\label{obs-exps}

We start with a simple observational study of a single agent interacting with its environment.
%AdamC: for flow and shortness reasons I commented out the following line
%Only once we have mastered the art of observation can we begin to design controlled experiments to understand the properties of our agents and their underlying learning systems.
%In an observational study, we---the experimenters---do not attempt to control any element of the study.
%A classic example is the ``bakeoff''-style of experiment, where several baseline algorithms are tested across several benchmark environments.
%The results of such a study lay the groundwork for identifying questions of interest for further experimentation.

Because much experimentation and observation in reinforcement learning occurs in deterministic computer simulations, we as researchers have a far greater degree of control in the design of our experiments than most natural sciences.
This control can often become distracting and cause us to lose sight of our original goals as empiricists.
As such, throughout this paper we will reference parallels to other fields of empirical science in order to provide grounding for our empirical practices and to build intuition.

\begin{myexample}
Imagine an animal learning laboratory where we study how quickly rats can learn to navigate a maze.
We place an individual rat in the maze and record some demographic details; its height, weight, age, sex, and so on.
We then observe the rat's behavior as it begins to explore its environment and learns to obtain a block of cheese at the end of the maze.
Throughout, we measure time-to-completion, the number of wrong turns taken, and some qualitative measures such as the perceived frustration of the rat throughout its learning process.

It is clear that some parameters of the individual rat will play a role in the time it takes for the rat to complete the maze.
Highly fit rats will likely complete the maze faster than particularly lazy rats, though the degree of influence is complex and unknown.
Further, these rats are all individuals of the same subspecies and differences between the individuals are unknown to the researcher (i.e., the researcher does not observe the genetic make-up of each rat, nor does the researcher know each rat's detailed history of experiences before entering the lab).
However, some elements of these rats are expected to remain consistent because they belong to the same subspecies---for example, this subspecies is known to be generally smaller and more docile than other subspecies of domesticated rats.
\end{myexample}

The field of reinforcement learning shares many similar motivations in both how and why we conduct experiments as in this example.
One such similarity is in how we measure an agent's performance.
However, this is only the first level of analysis.
The goal in animal learning is not to understand how quickly rats can complete mazes, but rather to use the maze task as a platform to probe the rat's ability to learn and explore.

The second analogy is in how we obtain repeated trials.
We can describe several characteristics of a particular subspecies of rat---one subspecies is more docile, another generally larger---and these characteristics inform us about general behavior of that subspecies.
However, within a subspecies there is still variation across individuals. For example, although lab rats are generally docile, we might still encounter a particularly ornery rat.
To understand and characterize an entire subspecies, the researcher often averages over these individuals.
Likewise, we are typically interested in characterizing the performance of an algorithm, which itself will be used to instantiate many individual agents.
%AdamC: for flow and shortness reasons I commented out the following line
%In order to ascribe behaviors to a learning algorithm, we must first account for the individual differences of the agents it produces.

Let us establish some terminology before diving into our first set of experiments.
Throughout this paper, we use the term \textbf{agent} to refer to a single entity interacting with its environment; analogous to an individual rat in a maze.
We use the term \textbf{algorithm} to refer to a process which produces a set of agents, both by specifying initial conditions such as the initial weights of a neural network, and by specifying the learning rules by which the agent adapts to observations.
Algorithms typically have configuration parameters---often called hyperparameters---which modify the set of individuals produced by the algorithm.
We call an algorithm with a particular hyperparameter configuration a \textbf{fully-specified algorithm}~\citep{jordan2020evaluating}, which is analogous to a particular subspecies of rat such as \emph{Rattus norvegicus domestica}, the common lab rat.
Finally, an \textbf{unspecified algorithm} refers to an algorithm where some (or all) hyperparameters have not been configured. This is analogous to a species of rat, such as \emph{Rattus norvegicus}.

\subsection{Experiment one: a demonstration}\label{sec_first}
% --> First step is forming a question. What do you want to study? Here is an example.
The first step towards designing an effective experiment is identifying the scientific question (or hypothesis) of interest.
In this first section, we will focus on the simpler observational studies which typically seek to answer questions of the form ``\emph{How well does algorithm A perform on environment E?}''
We will specifically investigate the Expected SARSA (or E-Sarsa) algorithm on a simple maze gridworld environment.
Details of the learning algorithm and environment can be found in Appendix~\ref{app_experiment-details}.
For now, it is sufficient to know that individual agents have sufficient learning capacity to efficiently find a near optimal policy in this environment.

% --> Second step is making experiment design choices. Here are a few examples
Now that we know which agents we want to observe and on what environment, there are still many design decisions to be made.
For example: (1) how many time steps will each trial or run contain? (2) if a terminal state is not reached after $n$ steps, will we artificially terminate the episode? (3) if the task is episodic, how many episodes should we run?
In general these choices should be made based on what you want to show. We will illustrate our thought process with an example experiment.

Let us start with a common and simple case: running E-Sarsa on our maze for a fixed budget of time-steps measuring online performance. The goal in this problem setting is to get high episodic returns, so the performance we report is the discounted return for an episode.\footnote{Many deep reinforcement learning algorithms are deployed with discounting even in undiscounted episodic cost-to-goal tasks. When this choice is made, it should be considered part of the internal mechanics of the agent: a hyperparameter inside the algorithm that induces additional discounting. This additional discount is not part of the environment and thus the performance metric should be the undiscounted return.}
We use a fixed budget of agent-environment interactions, meaning we run E-Sarsa on the maze for $n$ steps. If $n$ is large enough, then the agent will reach a terminal state and begin a new episode many times. With a fixed budget of steps, the agent will complete a variable number of episodes in $n$ steps depending on how good the agent in this problem setting.

We use a fixed budget of steps to better reflect that we care about online performance, where each sample matters. It lets us ask: after $n$ steps of environment interaction, how good is the agent's policy? If instead we use a fixed number of episodes, then some agents can get much more experience for learning. For example, a model-based agent could thoroughly explore in the first episode to learn its model, taking 9999 steps, whereas a model-free agent might find the goal in 100 steps. On the next episode, the model-based agent might already have a near optimal policy, looking like it learned much faster. In contrast, if we had considered the number of environment steps, then we might not find a big difference. It is not clear that our algorithms are designed to actually do this, but nonetheless we should measure what best reflects the goals of our experiment.

One secondary advantage of using a budget is that it avoids highly variable runtimes. Under a fixed budget of episodes, poorly performing agents can have very long episodes, resulting in much longer runtimes. See \citet{machado2018revisiting} for further discussion on this in the context of Atari.

Now let us consider how we can plot the learning curve for this agent, to see both how quickly it is learning as well as observe its final performance at the end of learning.
To plot the return at time step $t$, we use
the return for the current episode $G_j$ that began on timestep $j \ge 0$,
$$G_j \doteq R_{j+1} + \gamma R_{j+2} + \ldots + \gamma^{T-1} R_{j+T},$$
where $T$ is the (variable) length of the episode and $j \leq t \leq j+T$. The learning curve will be a piecewise linear step function, where the plotted performance will be the same for every step of the episode. For this particular environment, the rewards in this return are all zero except $R_{j+T} = 1$, making $G_j = \gamma^{T-1}$.

\begin{figure}
  \centering

    \includegraphics[width=0.5\textwidth]{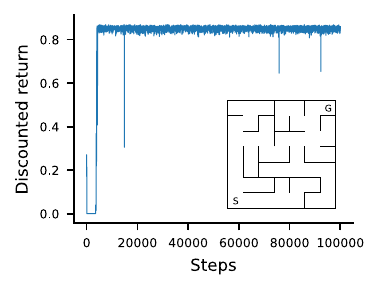}
  \caption{\label{fig:ES_PW_1R}
    A single run of an Expected Sarsa agent on a simple, tabular maze environment. The return rate for this agent is $\PerfRV = 0.827$. The agent has near optimal performance near the end of the curve, as it reliably reaches the goal in 15 to 17 steps, with the return hovering around $0.99^{17} = 0.84$ to $0.99^{15} = 0.86$.
  }
\end{figure}

Figure~\ref{fig:ES_PW_1R} shows the result of our first simple experiment. We plot the return per step of our E-Sarsa Agent on the maze. The curve starts just below 0.2, because the agent randomly found the goal early in learning in under 200 steps (notice $\gamma^{200} = 0.13$). Then it is flat at zero for some time as its next episode takes so long that the return is effectively zero. The longer the flat portion, the longer it took the agent to complete the episode. The curve has a step-profile because of the way we plot episodic return versus steps and because the data only reflects the performance of a single run of the experiment.

It will also be useful to be able to summarize this performance over time. In this case, because we care about online performance---and so how much reward the agent receives while learning---it is natural to report the average over the points in the curve, or the sum (often called the area under the curve). We call this average over the learning curve the \emph{return rate}.
Such summary or aggregate performance numbers $\PerfRV$ are particularly useful when we want to compare agents (Section \ref{sec_comparing}) or reason about quality of hyperparameters (Section \ref{sec_hypers}).

\begin{Remark}{}{measures_reference}
Other measures of performance and other aggregation functions for our learning curves can be considered. For simplicity in this work, we largely focus on measuring online episodic returns with this simple averaging aggregation. We discuss other choices in more depth in Appendix \ref{sec_metrics}.
\end{Remark}

We can see that this agent reached reasonable performance (near optimal) in this maze environment, though it is a bit hard to say if it learned quickly because we have no comparators. But, in any case, this was only one run! Perhaps we just got lucky? We need more independent evaluation of the agent to better characterize the performance.

\subsection{Experiment two: characterizing variations in performance}\label{sec_single_more}
Our first result was demonstrative in nature, however, most of the time we are interested in results that capture the reliability of our algorithms. Why would our algorithms perform differently if we ran them more than once?
There are two primary sources of variation across agents produced by an algorithm; think \emph{nature versus nurture}.
The first source of variation we encounter occurs when we initialize a particular agent.
Often, our function approximators---particularly neural networks---are randomly initialized, causing differences between agents even before data is observed.
The second source of variation comes from differences in the stream of data that a particular agent observes throughout its life (e.g., due to stochasticity in the environment or differences in actions selected by different agents).

\begin{figure}
\centering
\begin{subfigure}{.47\textwidth}
  \centering
  \includegraphics[width=0.9\linewidth]{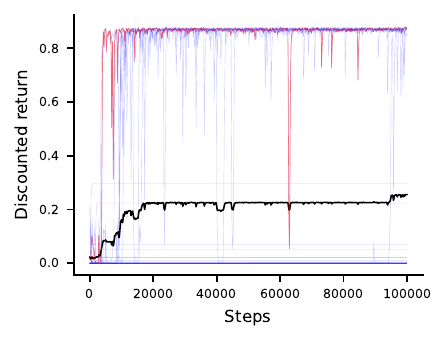}
  \vspace{-0.1cm}
  \caption{Performance of reinforcement learning agents in a maze.}
  \label{fig:all_runs_maze}
\end{subfigure}
\hfill
\begin{subfigure}{.47\textwidth}
  \centering
  \vspace{-0.1cm}
  \includegraphics[width=0.8\linewidth]{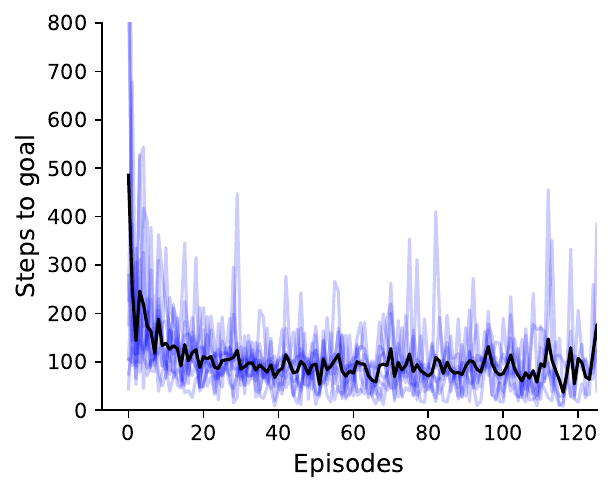}
  \vspace{0.1cm}
  \caption{Performance of real rats in a water maze}
  \label{fig:raw_rat_data}
\end{subfigure}
\caption{Understanding variability in agents. \textbf{(a)} 30 individual E-Sarsa agents on a simple, tabular maze environment. The thick black line shows the mean over individual agents over time. \textbf{(b)} Raw data from 10 rats running a water maze. }
\end{figure}

In Figure~\ref{fig:all_runs_maze} we plot the best and worst performance of the E-Sarsa agent in the Simple Maze environment, highlighting a large variation in performance. The right side Figure~\ref{fig:raw_rat_data} shows data from a real animal learning experiment: 10 different rats running a water maze on 10 different days.\footnote{
  It is common to exclude the data from poorly performing animals (after several retries). This practice has been adopted because scientists have identified genes that cause poor performance on particular tasks \cite{vorhees2006morris}.
  In reinforcement learning, we have no such prior knowledge which easily justifies removing data; dropping outliers should be done with caution.
} It is interesting how much variation we see in the rat data. It should not be too surprising to us that learning agents, even using the same algorithm (same species), could exhibit quite a bit of variability.
% compare to the variation we observe in our simple puddle world domain.

Our efforts as reinforcement learning empiricists are not that different than animal learning researchers and thus the motivations for repeated trials is also similar. We do not make claims about the abilities of a particular species of rat from an individual run through a maze. Each rat will be slightly different, due to random genetic factors and raising. Each time an individual is run through the maze things will be slightly different: lighting, humidity, the way the researcher holds the rat, {\em etc.} This maps pretty well onto the differences we see in agent initialization (e.g., neural network initialization) and environment variability (perhaps simulated by sticky actions in Atari \citep{machado2018revisiting} or real-world issues like motor actuations).

\subsection{Distributions matter!}

The variability in performance across agents gives rise to a distribution over performance for our algorithm.
If we run an algorithm on a single environment multiple times, then we are repeatedly sampling from a distribution: $\mathbb{P}(\PerfRV)$ where $M$ is the aggregate performance for a run.
The stochasticity comes from changing the algorithm's random seed for initialization and decision-making, and by changing the environment's random seed for simulated noise and start states.
If $\mathbb{P}$ is skewed or bimodal, then we might need more than a dozen independent runs to estimate the mean and variance. Worse, these simple sample statistics might be misleading.

\begin{SCfigure}[6][htb]
  \centering
  \includegraphics[width=0.34\textwidth]{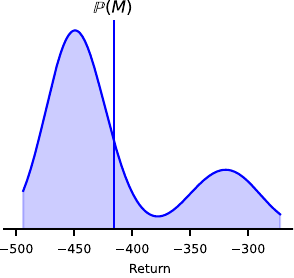}
  \caption{{\footnotesize \label{fig:perf-dist-demo} Performance distribution $\mathbb{P}(\PerfRV)$ of DQN on Mountain Car with stepsize=$2^{-9}$. The performance $M$ is the final performance of the agent after 100,000 steps of learning. The density around values for $M$ represents the probability of an experimental trial yielding a given outcome. For instance, if we run DQN for a single random seed we will most likely observe an agent that achieves a return of approximately -450 by the end of learning. This density is estimated using 1000 independent DQN agents using the \texttt{gaussian\_kde} kernel density estimation function in {\em scipy}.}}
\end{SCfigure}

In practice $\mathbb{P}(\PerfRV)$ can be skewed or multi-modal, as we can wee in Figure \ref{fig:perf-dist-demo}.
The sample mean (blue solid line) suggests the average agent would achieve a return of approximately -425.
However, a more complete description would be: any randomly selected agent is most likely going to receive a return of -450, while some agents will receive a return of -250.

\subsection{Reporting variability in performance}
\label{var-of-perf}
The performance distribution above provides a rich performance summary of many individual agents, compared to simply averaging learning curves.
These performance distributions, however, often require an infeasible amount of data to construct.
Instead, we can consider other descriptive summary statistics of this distribution beyond the mean.

One such pair of summary statistics are upper and lower percentiles, $(a,b)$, that reflect the range of performance.
Using percentiles, we can describe the range of performance values that, say, 90\% of observed agents achieve.
However, because we are not observing all possible future agents, we have some uncertainty whether this range accurately reflects 90\% of \emph{all possible} agents produced by this algorithm.

\textbf{Tolerance intervals} provide a distribution-agnostic way to summarize the range of an algorithm's performance while taking into account this uncertainty.
Like a confidence interval, we specify an acceptable level of uncertainty, $1-\alpha$ (e.g. $\alpha = 0.05$), then, unlike a confidence interval, we specify a proportion of individual agents captured by the range (e.g. $\beta=0.9$).
Constructing a $(\alpha, \beta)$-tolerance interval is rather straightforward; essentially, you compute the empirical upper and lower $(1-\beta)/2$ percentiles, then widen the interval based on the number of samples used.
Many standard computing packages include support for tolerance intervals and we provide more details in Appendix \ref{app_tolerance}.

\begin{figure}[htb]
\centering
\begin{subfigure}{.33\textwidth}
  \centering
  \includegraphics[width=\linewidth]{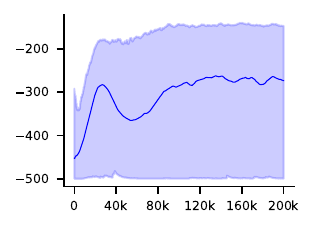}
  \caption{$\beta = 0.9$ with mean}
  \label{fig_maze_tolerance_one}
\end{subfigure}%
\begin{subfigure}{.33\textwidth}
  \centering
  \includegraphics[width=\linewidth]{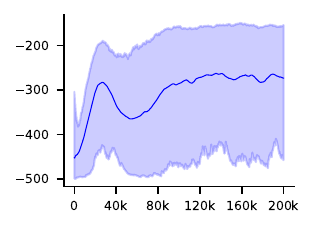}
  \caption{$\beta = 0.7$ with mean}
  \label{fig_maze_tolerance_two}
\end{subfigure}%
\begin{subfigure}{.33\textwidth}
  \centering
  \includegraphics[width=\linewidth]{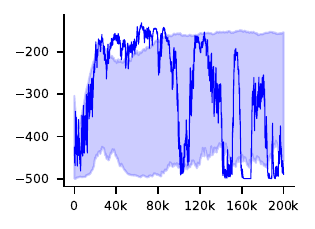}
  \caption{$\beta = 0.7$ with {\bf median}}
  \label{fig_maze_tolerance_three}
\end{subfigure}%
\caption{Tolerance intervals for the discounted return of DQN on Mountain Car, over 50 runs with $\alpha = 0.05$. Recall $\beta$ specifies the percentage of the distribution considered for the tolerance interval. \textbf{(a)} Tolerance interval with $\beta = 0.9$, with mean performance. \textbf{(b)} Tolerance interval with $\beta = 0.7$, with mean performance---notice the later is tighter. \textbf{(c)} Tolerance interval with $\beta = 0.7$, with {\bf median performance}.}
\vspace{-0.3cm}
\end{figure}

Let us return to our experiment with DQN on Mountain Car and examine the variability. We ran DQN for 50 runs, varying the seed for the agent and environment together.
We plot mean performance with an $(\alpha = 0.05, \beta = 0.9)$ tolerance interval in Figure \ref{fig_maze_tolerance_one}. The range in the plot is created by computing tolerance intervals for the 50 values at each time step $t$.

Finally, we might also want to visualize the performance of the \emph{median} agent instead of the \emph{mean} performance over agents. The mean performance does not correspond to any of the agents in the 50 runs. It might be useful to know how a representative (median) agent performed. We visualize this in Figure \ref{fig_maze_tolerance_three}. We take the average episodic return over learning in each run, to get $\PerfRV_1, \PerfRV_2, \ldots, \PerfRV_{50}$, and pick the run $j$ that is the median in $\PerfRV_1, \PerfRV_2, \ldots, \PerfRV_{50}$. We plot the learning curve for this agent. We can see that this median agent exhibits considerably higher variation per timestep than the mean learning curve over agents---{\bf this is not uncommon, actually!}
We also highlight that the tolerance interval is not centered around this median agent. Rather it is centered around the median over all agents taken at each timestep.

Standard deviations and tolerance intervals reflect the variation in performance. As we get more and more runs, the sample standard deviation and tolerance intervals approach their true values: the true standard deviation and the true probability interval that captures $\beta$ proportion of the population. They do not shrink to zero with more runs, unlike confidence intervals, which we discuss next.

\subsection{Reporting confidence in performance estimates}\label{sec_ci}
The majority of results in the reinforcement learning literature report confidence intervals which are very different from what we have discussed so far. Mean and variance plots and tolerance intervals attempt to capture the variation in the distribution of performance. As you collect more and more data---more runs---you get more and more accurate summaries of the variation. A confidence interval on the other hand has a different goal: to capture how certain we are in our estimate of some statistic about agent performance. As you collect more and more data we become more and more confident that our estimate has converged.

Confidence intervals allow us to report our uncertainty in the mean estimate, for example. We obtain a confidence interval $(l, u)$ for a given confidence level $1-\alpha$, where the interval is wider if we desire a higher confidence: wider for $\alpha = 0.01$ than $\alpha = 0.05$. The interpretation is that there is a low likelihood $\alpha$ that the true mean falls outside $(l,u)$. The uncertainty comes from the finite sample we observe, meaning our interval could have been different had we seen a different sample (different runs). In other words, we can say that at least $1-\alpha$ percentage of the time with a different random sample, our interval would contain the true mean. It is possible that we were unlucky and have the finite sample where the true mean is not in our interval.%, but it is unlikely that we have that interval.
\footnote{Mean performance across runs is commonly reported in reinforcement learning, and so we discuss appropriate empirical procedures here for estimating it. It is important, however, to reflect for yourself on when it is useful or not useful to use this evaluation. As we saw in the previous section, the mean performance is not necessarily reflective of any agent. The behavior of each agent can actually be quite erratic, but the mean performance might be reasonably smooth. In some settings, we want each agent to behave reasonably; reporting just the mean does not allow us to see potentially poor or erratic behavior in these agents. In other settings, such as when the environment is highly stochastic and the primary cause of variability, then it might be sensible to report just the mean performance.}

To make such probabilistic statements, we have to make different assumptions about the underlying distribution over performance.  The distribution could be Gaussian (normally distributed), but more generally could be any distribution, such as the bimodal, skewed distribution in Figure \ref{fig:perf-dist-demo}. We cannot know ahead of time exactly what our distribution looks like. But, there is a wealth of literature on selecting appropriate confidence interval approaches (see \citet{japkowicz2011evaluating} for a good reference).

A reasonable choice is to obtain a confidence interval using the Student t-distribution \citep{student1908probable}. This choice assumes the underlying distribution is approximately Gaussian. A typical recommendation before using this approach is to visualize the empirical distribution over your samples, to see if normality is a reasonable assumption---a so-called graphical method.\footnote{Alternatively, one could use a goodness of fit test like the Kolmogorov–Smirnov test or other statistical techniques. Many software packages such as Matlab, R, and SPSS have implementations you can use.}  For example, we could plot the 50 sampled scalars $\PerfRV_1, \PerfRV_2, \ldots, \PerfRV_{50}$ and see if they are concentrated around the mean value, or even use a package to plot an empirical distribution. More generically, if we assume we have $\numest$ samples of performance, the Student t-distribution confidence interval is of the form
\begin{align*}
\left[\overline{\PerfRV} - t_{\alpha, \numest} \frac{\hat{\sigma}}{\sqrt{\numest}}, \overline{\PerfRV} + t_{\alpha, \numest} \frac{\hat{\sigma}}{\sqrt{\numest}}\right] \quad\quad \text{where } \overline{\PerfRV} \defeq \frac{1}{\numest} \sum_{i=1}^\numest \PerfRV_i \quad \text{and }
  \hat{\sigma}^2 &\defeq\frac{1}{\numest-1} \sum_{i=1}^\numest (\PerfRV_i - \overline{\PerfRV})^2
  .
\end{align*}
The multiplier $t_{\alpha, \numest}$ depends on the confidence level and the number of samples. For example, for $\alpha = 0.05$, as the number of samples increases, $t_{\alpha, \numest}$ gets closer to the typical 1.96 for Gaussian distributions.\footnote{Using a Gaussian confidence interval requires knowing the true variance. Because we have to estimate it from data, we actually have uncertainty in this part of our interval as well. Therefore, our interval is actually a bit wider than if we knew the true variance. With more samples, our estimate of the true variance (true standard deviation) becomes accurate and so the multiplier approaches the value of the multiplier we get if we knew the true variance.} This multiplier can be obtained from the Student t-distribution table, or again computed using standard computing packages. As an example, for $\alpha = 0.05$, with $\numest = 3$ (three samples, two degrees of freedom) we have $t_{\alpha, \numest} = 4.303$, for $\numest = 10$ we have $t_{\alpha, \numest} = 2.262$ and for $\numest = 1000$ we have $t_{\alpha, \numest} = 1.962$.

The confidence interval itself shrinks with more samples, because the standard error term $\hat{\sigma}/{\sqrt{\numest}}$ goes to zero. In our setting, this means as we get more and more runs, our confidence interval around our mean estimator $\overline{\PerfRV}$ shrinks to zero until we can confidently claim that we have an accurate estimate of the mean.

\begin{figure}
\vspace{-0.8cm}
\centering
\begin{subfigure}{.36\textwidth}
  \centering
  \includegraphics[width=\linewidth]{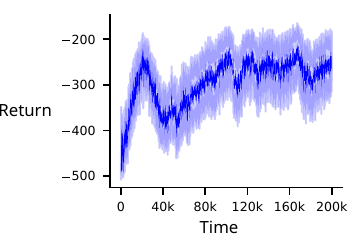}
  \caption{$\alpha = 0.05$ with Student's t}
  \label{fig_maze_ci_one}
\end{subfigure}%
\begin{subfigure}{.31\textwidth}
  \centering
  \includegraphics[width=\linewidth]{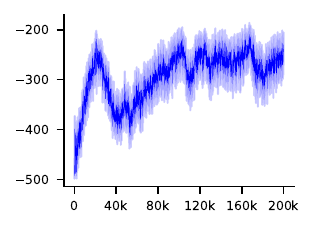}
  \caption{$\alpha = 0.3$ with Student's t}
  \label{fig_maze_ci_two}
\end{subfigure}%
\begin{subfigure}{.31\textwidth}
  \centering
  \includegraphics[width=\linewidth]{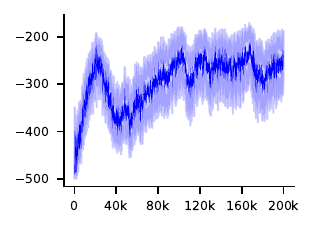}
  \caption{$\alpha = 0.05$ with bootstrap}
  \label{fig_maze_ci_three}
\end{subfigure}%
\caption{Confidence intervals around the discounted return of DQN on Mountain Car averaged over 30 runs. \textbf{(a)} Student's t-distribution confidence interval with $\alpha = 0.05$.  \textbf{(b)} Student's t-distribution confidence interval with $\alpha = 0.3$. \textbf{(c)} Bootstrap confidence interval with $\alpha = 0.05$, with mean performance.}\label{fig_ci_dqn}
\vspace{-0.3cm}
\end{figure}

In our our third experiment, we compute and plot a confidence interval around the mean estimator---instead of a tolerance interval---using the same data as in the last section. We can see in Figure \ref{fig_maze_ci_one} that the confidence interval with $\alpha = 0.05$ is already quite tight, for these 50 runs. This shaded region reflects our uncertainty in our estimate of the mean, whereas the tolerance interval in Figure \ref{fig_maze_tolerance_one} reflects the variation around the mean (and so is wider and does not shrink to zero).

We can ask how many samples $\numest$ (or runs) we need before it is reasonable to compute this confidence interval. We can actually obtain a statistically valid confidence interval, even with only two samples! The interval itself will simply be wider, because $t_{\alpha, \numest}$ will be larger as will the standard error. However, it is difficult to gauge, with only two samples, if it is appropriate to make the assumption that the underlying distribution is Gaussian. You need to obtain enough runs to decide if the Student t-distribution confidence interval is appropriate.

In many cases we may not be confident that our performance distribution is Gaussian, or we might even believe it is not Gaussian. In that setting,
a natural alternative to the Student's t confidence interval is to use bootstrap-based statistics to generate confidence intervals.
The bootstrap procedure is simple, but not the same as boostrapping in TD learning. As before, we get $\numest$ measures of performance for an algorithm, one for each of the $\numest$ random seeds. We then generate a new dataset by resampling $\numest$ values from the original dataset \emph{with replacement}. Finally, we repeat this resampling procedure for a total of $\numrepeats$ times, with $\numrepeats$ usually very large (e.g. $\numrepeats = 10,000$).
For each of these new datasets, we compute the statistic of interest (e.g. the mean) then measure the variability of that statistic over all $\numrepeats$ datasets.
To create a 95\% confidence interval, we report the $0.025$ percentile of the estimated statistic over all $\numrepeats$ estimates as a lower-bound and the $0.975$ percentile as an upper-bound.

A major advantage of bootstrap-based methods is that they often require very few assumptions about the underlying data.
This advantage comes at a cost, however, because bootstrap methods generally require more data points to provide tight confidence intervals.
For this reason, our recommendation is to default to bootstrap-based methods for most comparisons, but to check the underlying distributions to see if more powerful methods---such as Student t-distribution confidence intervals---can apply without breaking assumptions. We visualize the bootstrap confidence interval using the above procedure in Figure \ref{fig_maze_ci_three}.

\begin{Remark}{}{firstsummary}
We did not seem to have to be so careful with tolerance intervals. But there too we had to account for uncertainty in our percentile estimates. The same distributional question arises. If we know we have an underlying Gaussian distribution, then we can get a better estimate of these percentiles with fewer runs. It is common, however, to default to distribution-free tolerance interval calculations, that make few assumptions about the data.
We discuss this further in Appendix \ref{app_tolerance}.
\end{Remark}

\subsection{Do we really need more runs?}\label{sec_moreruns}
Nobody wants to run their experiment longer than needed. Just increasing the number of runs makes our experiments take longer, and has real environmental costs. Ideally, one would hope that advanced statistical tools can save us. Unfortunately, there is an inherent trade-off between making some assumptions and having tighter confidence intervals, and avoiding assumptions and having potentially useless confidence intervals. In this section, we highlight that we may need to do more runs, especially for the types of agents that we currently analyze because they can produce highly skewed performance distributions.

\begin{figure}[t]
\centering
\begin{subfigure}{0.48\textwidth}
  \includegraphics[width=\textwidth]{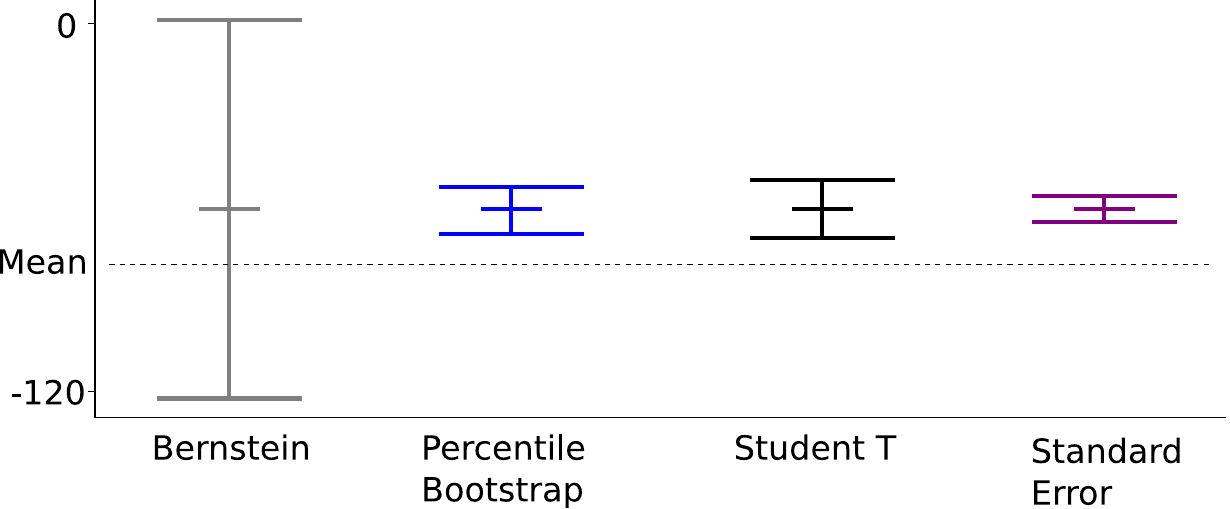}
  \caption{Confidence intervals with 10 runs.}
\end{subfigure}
\hfill
\begin{subfigure}{0.48\textwidth}
  \includegraphics[width=\textwidth]{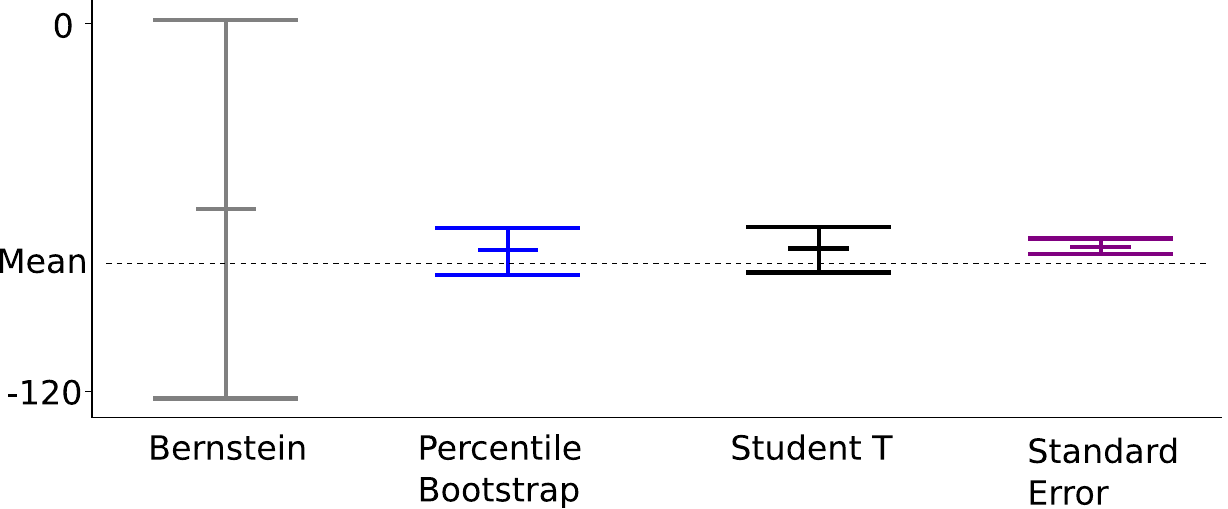}
  \caption{Confidence intervals with 50 runs.}
\end{subfigure}
\caption{\label{fig:skew_confidence_intervals}
  Confidence intervals about the mean total discounted return of DQN on the PuddleWorld environment.
  Long-tail behavior in the performance distribution violates assumptions for the Student's t-distribution, whereas the percentile bootstrap fails when too few samples are used.
  Increasing the number of samples---right subfigure (b)---can resolve the issue for the percentile bootstrap. All of these intervals should be 95\% confidence intervals, except for the standard error. We include the standard error only to highlight that it can be much narrower than the corresponding 95\% confidence interval.In this instance, for $n = 10$, the standard error represents a Student-t confidence interval with $\alpha = 0.5$ (because $t_{\alpha=0.5, \numest = 10} = 1$) and for $n = 50$, we have $\alpha = 0.3$ ($t_{\alpha=0.3, \numest = 50} = 1$).
}
\end{figure}

Consider the following example.
Imagine we have collected 10 runs of the DQN algorithm on the PuddleWorld environment and we wish to report DQN's average performance. We can compute a confidence interval around this mean estimate, to reflect our uncertainty due to only having 10 runs. There are several assumptions we could make, resulting in different confidence intervals. We visualize several options in Figure~\ref{fig:skew_confidence_intervals}. In this synthetic example, we can actually compute the true mean and check: did our confidence intervals capture the mean? We can see the estimated average performance for these 10 runs is far from the true average performance of DQN in this environment.
Further, four of the confidence intervals provide overly optimistic ranges and fail to capture our estimate of true mean (from 250 samples).

The sample Bernstein confidence interval does reasonably capture the population mean in Figure~\ref{fig:skew_confidence_intervals}; a natural conclusion might be to prefer such confidence intervals which make minimal distributional assumptions.
However, the Bernstein bound is highly conservative meaning provided confidence intervals are incredibly wide even for large numbers of runs.
Unfortunately, this inhibits truly understanding whether we have accurately captured the average performance---in the case of PuddleWorld, the confidence region covers a majority of the range of possible means---requiring substantially more runs to present statistically meaningful results.
In the example above, we required as many as 1000 runs to detect differences between DQN and an alternative algorithm using the sample Bernstein confidence interval, while the percentile bootstrap required only 30 runs.
\newpage
\begin{wrapfigure}[8]{l}{0.3\textwidth}
\vspace{-0.4cm}
  \centering
  \includegraphics[width=\textwidth]{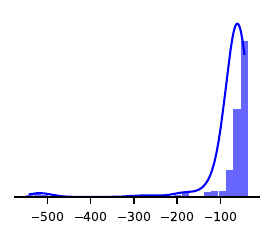}
\end{wrapfigure}
Unfortunately, these examples where confidence intervals fail are far from rare.
To see why they fail in this instance, we need only look at the performance distribution of DQN on PuddleWorld (see inset figure to the left).
The distribution is long-tailed with some very low performing runs occurring with low probability (approximately 5\% of the time).
In this case, all 10 of our samples were high performing causing us to overestimate the average performance while simultaneously underestimating the variation in performance.
Note the probability of receiving 10 samples near the right-side mode is approximately 60\%---not at all uncommon.
The only way to resolve this issue is simply to collect more runs of DQN.
For this specific example, we found that approximately 20 runs were sufficient to accurately reflect the high variation and obtain accurate bootstrap confidence intervals, and 30 runs were sufficient to accurately estimate the average performance. This provides a clear example of how dangerous it can be to declare the variance is low based on a few runs. Generally, we may need a large number of runs to compare agents \citep{jordan2024position}, which we discuss more in Section \ref{sec_comparing}.

One proposed solution to allow for a smaller number of runs, knowing that we have such distributions, is to instead report an estimate of the interquartile-mean (IQM) \citep{agarwal2021deep}. The IQM takes the mean of the points in the interquartile range, namely between the 0.25 and 0.75 percentiles. This statistic is more robust to outliers, because these potentially larger magnitude values are not included in the mean calculation. The proposed estimate of the IQM is to drop the 25\% highest samples (runs) and 25\% lowest samples before computing the mean of the remaining 50\% of the data \citep{agarwal2021deep}.

As with any choice, we need to be cautious about whether the goal was to estimate the IQM, or whether we chose it because it provides a convenient way to reduce the number of runs. In some cases, it is a useful statistic, with the added benefit of being a robust statistic. In other cases, it may not capture key properties of the algorithm that we care about.
Take the DQN algorithm as example. We observed low-probability catastrophic failure events for DQN across nearly every environment we tested.
In Lunar Lander, some agents would simply fly off into oblivion, obtaining incredible amounts of negative reward until the episode was mercifully terminated due to episode cutoffs.
In Cliff World, some DQN agents would get stuck in a corner perpetually in every single episode, never learning to find the goal. Even worse, a small subset of agents would learn to always jump into the cliff immediately and obtain massive negative rewards.
In this case, removing these outlier agents---the agents whose performance do not conform to our pre-existing notions of how DQN \emph{should} behave on simple environments---is not helping to create a clearer picture of our algorithm, as we are simply ignoring its shortcomings.

Generally, IQM is most usefully applied across collections of environments. In Atari, some games are nearly trivial (e.g., Pong) and others are nearly impossible without prior knowledge (e.g., Montezuma's Revenge). Performance on trivial and impossible games can greatly skew performance measures, making it difficult to compare different agents. In such cases IQM makes a lot of sense and was the intention behind the origin design ({\em personal communication with authors}).

\subsection{Deciding on the length of an experiment}\label{sec_steps}

An important empirical choice for understanding your agent is the number of steps of interaction. If you choose a smaller number of steps, then you are evaluating early learning performance. If you choose a larger number of steps, then you are evaluating if the agent can reach near-optimal performance within a reasonable number of samples. If you choose a very large number of steps, then you may be evaluating if your agent can reach near-optimal performance and stably remain at this performance.

\begin{figure}[t]
  \centering
  \begin{subfigure}{.32\textwidth}
    \includegraphics[width=0.95\linewidth]{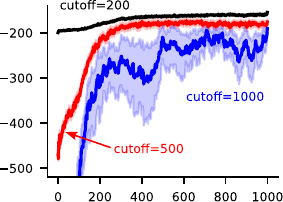}
    \caption{\label{fig:cutoff_means}
      Mean performance.
    }
  \end{subfigure}
  \begin{subfigure}{.32\textwidth}
    \includegraphics[width=0.95\linewidth]{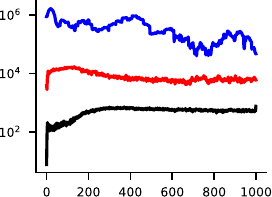}
    \caption{\label{fig:cutoff_variance}
      Variance over runs.
    }
  \end{subfigure}
  \begin{subfigure}{.32\textwidth}
    \includegraphics[width=0.95\linewidth]{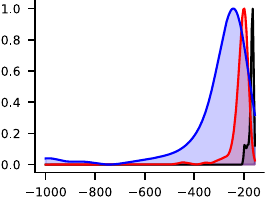}
    \caption{\label{fig:cutoff_dist}
      Distribution over runs.
    }
  \end{subfigure}
  \caption{\label{fig:cutoffs}
    Performance of the EQRC algorithm in Mountain Car for three different episode cutoff lengths: $\tau \in \{200, 500, 10000\}$.
    Smaller cutoffs clip the impact of outlier performance values, causing variance to be significantly under-reported and mean performance to be biased toward higher performance.
      %\vspace{-1cm}
  }
\end{figure}

One way to gauge if the number of steps reflects early learning or final performance is to examine the learning curve itself. If the curve is still increasing, then likely you are seeing early learning. If it has flattened for many steps, then you are seeing the final performance of the agent. To appropriately pick the number, you can run the agent for longer to gauge if it eventually levels off. Then you can report performance for a smaller number of steps to focus on early learning.

An additional choice related to steps of interaction is artificial episode cutoffs. Cutoffs are used to prevent an episode from becoming too long. An agent may get stuck, never finding its way to the goal and so never terminate the episode. A cutoff involves teleporting the agent after \verb+max_steps+ back to the start state, without entering a \emph{terminal state}. It is like picking up a robot that got stuck in a corner, and moving it back into a state from which it can learn.

For this reason, cutoffs can interact with exploration, and in particular very aggressive short cutoffs may misrepresent the performance of agents that get frequently stuck.
For example, in original implementations of Mountain Car, no early cutoffs were used and tile-coded agents would run for a few thousand steps in the first episode on average \citep{sutton2018reinforcement}.
In later implementations, an aggressive episode cutoff of 200 steps was introduced \citep{brockman2016openai}.
We show in Figure~\ref{fig:cutoffs} that performance for an algorithm that uses neural networks---called EQRC \citep{patterson2022generalized}---is much better with more aggressive cutoffs, with much lower variability over runs.\footnote{This change in default implementation illustrates the practice of problem adaptation---making an easier version of the problem to alleviate challenges in modern algorithms, such as neural networks struggling to learn from flat reward signals.}

We might choose to run the agent for a fixed number of steps, rather than a fixed number of episodes. In that case, we do not need episode cutoffs and experiments will terminate in a reasonable amount of time. Cutoffs can, however, be useful for other reasons. The main reason is that, due to stochasticity in environments, many agents may have some interactions that lead the agent to get stuck---effectively bad luck. This issue may not only have to do with the agent, but also potentially with the environment. We can set the episode cutoff to a large number, that is less than the number of learning steps, to avoid having runs where the agent is stuck in one episode forever. With a large cutoff, we are much less likely to make the problem too easy and can avoid significantly skewing the results. But, we obtain some of the reduced variance in runs to facilitate evaluation.

%\subsection{Summary}
\paragraph{Summary:}
We conclude each section with a summary of key actionable advice. Here, we summarize key points to consider when investigating an algorithm in one environment, assuming we have already specified the hyperparameters.

\begin{summary}[Key insights: evaluating a fully-specified algorithm]{}
\begin{enumerate}[leftmargin=7pt]
\setlength\itemsep{.1em}
\item A single agent's performance is informative, however, consider plotting the performance of individual runs to gain additional understanding of your algorithm.
\item Agent behavior, with the same algorithm and environment, can be very different across random seeds. Consider reporting this variability using sample standard deviations or tolerance intervals, rather than just reporting the mean or median performance.
\item Confidence intervals around the mean primarily tell us about our certainty in our mean estimate, not about the variability in the underlying agents. If you primarily care about understanding mean performance, then report means with confidence intervals. If you care about the behavior of each agent, not just the mean, then consider reporting tolerance intervals.
\item {\bf In general we advocate that you do \emph{not} report standard errors.} They are like a low-confidence confidence interval, and it is more sensible to decide on the confidence interval you want to report.
\item The required number of runs depends on your performance distribution, which is unknown to you. It is clear that in almost all cases 5 runs is insufficient to make strong claims, but even 30 runs can be insufficient if the distribution is heavily skewed.
\item Consider reporting {\em performance} versus {\em steps} of Interaction, rather than {\em performance} versus {\em episodes}. This choice ensures every agent receives the same number of samples every run. (See Section~\ref{sec_first})
\item Choose the number of steps of interaction to reflect early learning or ability to learn the optimal policy, rather than inheriting what was done in previous work.
\item Aggressive episode cutoffs can significantly skew results.
\end{enumerate}
\end{summary}

There are many things to worry about just to run a good observational study of single agent. But let's gain some perspective here: why should this be easy? It's not easy for our animal learning scientist to conduct and document the study of a few rats in a maze. They have to worry about so many other things that we do not: how to physically handle the animals, if their personal scent impacts the animals, uncontrolled genetic variations, running in real time, extra dull rats, and so on. In contrast, we can carefully enumerate our sources of bias, programatically vary conditions, exercise perfectly repeatable interventions, control almost all relevant sources of variation, perfectly record all relevant information, and run millions of experiments hundreds of times faster than real-time. %Scientists examining the real-world cannot cut corners, even though their empirical setting is more onerous than ours. Similarly,
If we truly want to understand our algorithms and gather sufficient evidence for our claims, then we need to take the scientific enterprise seriously. Science is first and foremost about understanding, not picking winners and losers.

\biblio

\section{Dealing with Hyperparameters}\label{sec_hypers}

% Algorithm: specification of learning rule, inputs, outputs, function approximation scheme, learned parameters and tuneable hyper-parameters. Classic example is tabular Sarsa(0) for control in Sutton and Barto
% Fully specified algorithm: all the same things as an algorithmic but also specifying how the function approximator and any learned parameters are initialized and all hyperparameters are specified. Different choices of hyperparameters for a given algorithm result in a family of related fully specified algorithms.
% Agent: an instance of a fully specified algorithm applied deployed in a particular environment. Given a fully specified algorithm, two agents may behave very differently due to randomness in the environment or randomness caused by initialization of the agent.

Almost all algorithms have hyperparameters. These are scalars that have to be selected by a person before running an experiment.
Typical hyperparameters in reinforcement learning include stepsizes and other optimization parameters like momentum and batch sizes; the eligibility trace parameter or the horizon for n-step methods; the target net refresh rate; and even the function approximation architectures used, which themselves can have many different hyperparameters (e.g., depth, number of nodes per layer, activation function, etc).\footnote{
It is reasonable to expect that some of these hyperparameters---like any learnable parameters---should adapt with time. However, we can consider hyperparameter adaptation as part of the algorithm; our job, as empiricists, is to specify the hyper-hyperparameters for that adaptive algorithm. For this reason, we define hyperparameters as the initial values set at the beginning of the experiment.}

The possible combinations of hyperparameters can be overwhelming. It is hard enough to properly evaluate an algorithm for a single hyperparameter combination, let alone having to consider this combinatorial space of algorithms. But, we can overcome this panic by stepping back and clarifying the goal of our experiment.
There are three typical goals: (1) understand the \emph{behavior of an algorithm across settings of the hyperparameters}, (2) optimizing hyperparameters to study an algorithm's \emph{idealized maximum performance}, and (3) \emph{simulating a  deployment scenario}.
As a field, we generally have more understanding about how to study hyperparameter sensitivity (corresponding to setting 1), though of course empirical design here is also nuanced; we discuss this in Section \ref{sec_hypersens}.

The second setting falls into the category of competitive machine learning.
Given the ability to extensively tune the performance of a learning system, what is the maximum capability we can hope to achieve?
Such studies are limited to specific problem settings, often called benchmark problems.
Unfortunately, there are multiple challenges that arise in this setting.
The primary challenge is statistical: it is hard to estimate the maximum value of a stochastic function, such as the maximum performance of a reinforcement learning system.
Another challenge arises in making fair comparison to baselines in the competition, it is difficult to ensure equal tuning effort is given to each competitor learning system.
We discuss this issue further in Section~\ref{sec_hyperopt}.

The final setting poses the greatest challenge.
Unlike supervised learning, we do not have a general purpose approach to select hyperparameters.
In supervised learning, cross-validation can be used with almost any algorithm to select hyperparameters according to generalization performance.
It is not obvious how to use cross-validation in reinforcement learning (see a more in-depth discussion in Appendix~\ref{app_cv}).
An additional challenge comes from the variety of use cases for reinforcement learning, such as solving simulated problems versus interacting with the real-world; having access to lots of data but limited compute versus lots of compute and limited data; or settings where taking exploratory actions is safe versus unsafe.
Each combination of these (and other) factors will lead to different methodologies to select hyperparameters for a reinforcement learning algorithm.
Understanding the performance of an algorithm in deployment depends not only on the algorithm and the hyperparameters tested, but also on the deployment scenario itself.
We discuss this issue further in Section~\ref{sec_hyperdeploy}.

\subsection{Understanding Hyperparameter Sensitivity}\label{sec_hypersens}

The goal of hyperparameter sensitivity analysis is to help us understand our algorithms. {\em This is not about optimizing hyperparameters to support SOTA claims!}
These insights can help identify serious sensitivities that suggest improvements to the algorithm are needed, they can help us understand changes in behavior as we interpolate across a space of different algorithms, or they can help identify hyperparameters which require joint tuning in order to provide good performance.
As a classic example, the trace parameter $\lambda$ in TD($\lambda$) algorithms interpolates between Monte Carlo algorithms as $\lambda \to 1$ and the original TD algorithm as $\lambda \to 0$.
A sensitivity study may reveal that $\lambda=0.9$ is an optimal choice for an environment, however, this is only one useful piece of information that we can derive from the study.
We may learn that performance becomes highly variable as $\lambda \to 0$ or that $\lambda$ near 1 diverges.
We might additionally learn that the performance suddenly drops off outside a narrow range of $\lambda$ suggesting that this algorithm will be difficult to tune in a new environment.

In order to evaluate our algorithm with different values of a hyperparameter, we need to collect enough data to provide reasonable estimates of our statistic of interest.
Say we wish to report the average performance across agents produced by our algorithm for each hyperparameter value, then for every hyperparameter setting we need enough agents to actually estimate that average.
This is no different than Section~\ref{sec_first} where we required multiple agents to evaluate an algorithm, except now we are evaluating multiple fully-specified algorithms; one for each setting of the hyperparameter of interest.
In the most basic setting, this means we need $N$ runs for every hyperparameter setting, $H$, for a total of $N \times H$ runs.
Clearly, this can become expensive quickly!

\begin{figure}[H]
  \vspace{-0.1cm}
  \hspace{-1.5cm}
  \floatbox[{\capbeside\thisfloatsetup{capbesideposition={right,center},capbesidewidth=8.9cm}}]{figure}[\FBwidth]
  {\caption{\label{fig:good_sensitivity}
    A good sensitivity curve that captures a wide range of the variable of interest and illustrates that performance changes smoothly as the hyperparameter changes.
  }}
  {\includegraphics[width=0.8\linewidth]{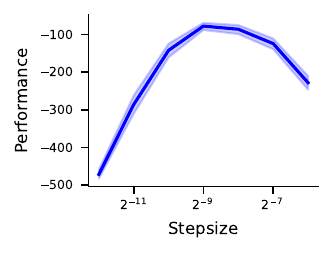}}
  \vspace{-0.5cm}
\end{figure}

\begin{figure}[H]
  \hspace{-1.5cm}
  \floatbox[{\capbeside\thisfloatsetup{capbesideposition={right,center},capbesidewidth=8.9cm}}]{figure}[\FBwidth]
  {\caption{\label{fig:pointy_sensitivity}
    A sensitivity curve where the range of tested values may be too wide, instead of being focused in the region of interest.
    We lose some information around the peak performance and the algorithm appears quite sensitive. This sensitivity might be an artifact of the plot---testing insufficiently many values---rather than a property of the algorithm.
  }}
  {\includegraphics[width=0.8\linewidth]{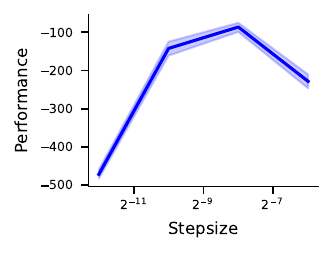}}
  \vspace{-0.5cm}
\end{figure}

\begin{figure}[H]
  \hspace{-1.5cm}
  \floatbox[{\capbeside\thisfloatsetup{capbesideposition={right,center},capbesidewidth=8.9cm}}]{figure}[\FBwidth]
    {\caption{\label{fig:bad_range}
      A sensitivity curve where we potentially missed the best performance.
      The best performing hyperparameter may be outside the range or may be the endpoint of the range, but we cannot tell with the information as available.
    }}
    {\includegraphics[width=0.8\linewidth]{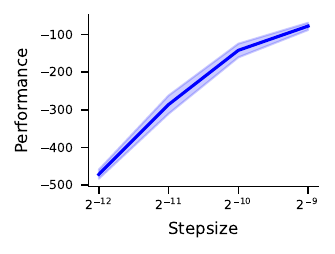}}
  \vspace{-0.5cm}
\end{figure}

\paragraph{Dealing with a single hyperparameter.}
Once we have obtained an estimate of performance for each setting of our hyperparameter, we can summarize the performance of this partially-specified algorithm as in Figure~\ref{fig:good_sensitivity}.
To create such a plot, we must first decide on a range for the hyperparameter then specify how intermediate values are selected within that range.
A common range for stepsizes is to use powers of $2$, to systematically cover the space.
% MARTHA: Should we keetp this?
%\footnote{
%  This log-uniform coverage of stepsizes arises from the fact that the learning rule often forms an exponential moving average parameterized by the stepsize.
%  Many hyperparameters have similar exponential effects such as the momentum parameter, the discount parameter $\gamma$, or the trace parameter $\lambda$.
%}
If there is a clear bowl or U-shape to the resulting curve---as in Figure~\ref{fig:good_sensitivity}---then this selection scheme was likely appropriate---though the curve is not always U-shaped.
If we observe sharp changes in performance---like the V-shape in Figure~\ref{fig:pointy_sensitivity}---you may need to sample more densely within that region to better understand the range of appropriate values for the hyperparameter.
These sharp changes often occur when the initial range of the hyperparameter is too large or when the distribution of tested values are concentrated around a region of poor performance---that is you missed the good ones.
If the best performance is at the one end of the range---like in Figure \ref{fig:bad_range}--- then this suggests the range was too narrow and may need to be systematically expanded.

Now that we have our sensitivity curve, how do we interpret it?
If the sensitivity curve is reasonably flat---the minimum performance is close to the maximum---for a wide range of hyperparameter values, then we might say that this partially-specified algorithm is insensitive and so it will not be challenging to define a fully-specified algorithm for deployment.
If the sensitivity curve indicates a large difference in performance within a narrow region of hyperparameter values, then we would say this partially-specified algorithm is highly sensitive and conclude that defining a fully-specified version for deployment might be difficult.

Such comprehensive experiments can be expensive, and it can be tempting to test only a small number of hyperparameter settings. However, we always have to ask ourselves if we have compromised our empirical design. {\bf There is no point in running a flawed experiment, even if it is more feasible in terms of computation.}
An alternative choice would be to ask research questions that better match our computational resources.
Remember, some empirical questions are simply more challenging to answer than others.
For example, it is challenging to convincingly show that an algorithm is insensitive---this would require testing a wide-range of values with a large number of intermediate values, possibly with multiple environments.\footnote{We could more thoughtfully sample the hyperparameters to get an accurate sensitivity curve. One could imagine an automatic procedure that starts broad, and iteratively samples hyperparameter choices between two where the values are very different, or focuses on regions of higher performance where we want the curve to be more accurate ~\citep{he2021automl,parker2022automated,eimer2023hyperparameters}. Bayesian optimization approaches provide such a strategy to find the best hyperparameters, and so the ideas there could potentially be adapted for this goal. Nonetheless, such an algorithm would still require testing a large number of hyperparameters.} Similarly, it is difficult to show empirically that an algorithm is divergent.

\begin{myexample}[Does TD diverge?]\label{example_bairds}
  We know that TD can diverge under off-policy updating.
  In fact, we can prove that TD always diverges in very specific scenarios such as Baird's counterexample \citep{baird1995residual}; though such theoretical characterizations may not always be possible and it is in these scenarios that we often rely on empirical evidence.

  For example, imagine you combine TD's update rule with a momentum term. Does this still diverge on Baird's counterexample? If there is no existing theory about the convergence of TD with momentum, you might turn to experiments to obtain some insights.
  Likely, we would start with the default hyperparameters for momentum (say setting the momentum term $\beta=0.9$).
  We find that this algorithm diverges!
  We test again with a different random seed and observe divergence; and again for another random seed, and again, and again\ldots
  After many such trials, we might conclude with high confidence that TD with momentum always diverges on this environment.
  This conclusion, however, is limited by the choice $\beta=0.9$. Does this conclusion hold for $\beta=0.99$ or $\beta=0.1$?

  To provide evidence that TD with momentum does not converge in this environment---regardless of hyperparameter setting---we must carefully sweep a dense set of values of $\beta$ and show divergence for every tested value.
  Without theory---or an infinite number of runs---we can never know with absolute certainty that TD with momentum always diverges.
  It is always possible that some untested random seed presents the data in exactly the right order such that TD with momentum converges, or possibly there is some perfect value of $\beta$ where TD converges in this environment.
  With empiricism, we accumulate a body of evidence which supports our claim; the more random seeds we test and the more hyperparameter values we sweep, the more convincing our body of evidence.
\end{myexample}

\paragraph{Assessing overall hyperparameter sensitivity}
\label{hyper_overall}
One of the primary goals of hyperparameter sensitivity studies is to understand how to improve our algorithms and develop those that are easier to tune.
In addition to explicitly visualizing performance versus a specific hyperparameter, we can also attempt to assess overall how sensitive an algorithm is to its hyperparameters. One approach \citep{jordan2020evaluating} is to treat all hyperparameters as unknown values. We model these unknowns as random variables and draw sample hyperparameter configurations.
For example, we might treat the stepsize, $\alpha$, as an unknown value and sample $\alpha \sim \text{Uniform}(0, 1)$.
Each sampled $\alpha$ produces a fully-specified algorithm, for which we can obtain a measurement of performance.
The amount that the performance changes as we change the hyperparameters then provides a measurement for the sensitivity of the under-specified algorithm on a given problem setting.

This procedure is akin to performing a sensitivity study across all hyperparameters simultaneously. Because the space of hyperparameters is combinatorial, exhaustively sweeping across configurations to obtain a measure of sensitivity is typically impractical.
Instead, this approach randomly samples points in the combinatorial space in order to compute the variation in performance over that space.
This presents a tradeoff: we can carefully and systematically investigate a small number of hyperparameters at a time, or we can try to investigate the entire space of hyperparameters with much less detail.
%Naturally, we can also use a mixture, where we systematically vary one hyperparameter, while randomizing over the rest.

\begin{Remark}{}{secondsummary}
Dealing with hyperparameters is an active area of research and thus our treatment of it here is necessarily limited. %Things become complicated and more computationally burdensome with more than one hyperparameter.
The interested reader can jump to Appendix \ref{app_adv_hyper} for an expanded discussion on handling multiple hyperparameters.
\end{Remark}

\subsection{Reporting Idealized Performance}\label{sec_hyperopt}

When we first introduce an algorithm, we may want to know how well it \emph{can} perform, on an environment. If it performs poorly on an environment, even with extensively tuned hyperparameters for that environment, then there may be an issue with the algorithm. Further, the algorithm may be conceptually appealing, but have new hyperparameters that have not yet been well-studied. If an argument can be made that adaptive algorithms could be developed for these new hyperparameters, then it can be appropriate to evaluate the behavior of the algorithm under nearly-optimal hyperparameter settings.

Though reporting performance (or behavior) for nearly-optimal hyperparameters is common in reinforcement learning, it has several serious pitfalls that you need to consider if you take this route.
%There are two primary difficulties in tuning hyperparameters for maximum performance in reinforcement learning.
The first difficulty is that estimating the maximum of a stochastic function is challenging and often requires a very large number of samples.
In our experiments, this translates to needing many random seeds and significant computational resources.
The second is that many reinforcement learning algorithms are notoriously sensitive to their hyperparameters and often contain dozens of hyperparameters to tune---recent algorithms make use of dozens of hyperparameters, not including those for the function approximator \citep{adkins2024a}.
Finally, we have to be careful about comparing two algorithms under this idealized setting where hyperparameters are optimized. If we allow one algorithm to have more hyperparameter settings, then differences in performance can be due to maximizing over more hyperparameter settings rather than differences in the algorithm. We discuss these issues in more details below.
% MARTHAC: These alst two sentences seemed out of place here, for this topic. I wonder if they should go at the beginning of this section? I put some of it above
%As a result, tuning a reinforcement learning algorithm can feel akin to alchemy---if we sprinkle in a bit of this and adjust a bit of that, then perhaps we can create gold!
%Designing new algorithms which retain similar levels of performance with less tuning is a worthy pursuit.

\paragraph{Maximization bias.}\label{sec_maxbias}
Several statistical challenges present themselves when tuning hyperparameters for maximum performance.
One challenge is \emph{maximization bias}, which is a form of statistical bias that can arise when estimating a quantity of the form $\max_h \CE{G}{h}$.
In our case, $h$ represents a hyperparameter configuration and $G$ represents the performance of our algorithm for a given configuration.
Because we do not know $\CE{G}{h}$, the average performance of our algorithm for a given hyperparameter configuration, we estimate it with samples, $\bar{G}_h \approx \CE{G}{h}$.
Then we estimate $\max_h \CE{G}{h}$ using $\max_h \bar{G}_h$. This estimate if prone to maximization bias, because $\mathbb{E}[\max_h \bar{G}_h] \ge \max_h \mathbb{E}[\bar{G}_h] = \max_h \CE{G}{h}$.
%This approximation will overestimate or underestimate the average performance for each hyperparameter setting.
%Finally, we evaluate each estimate $\CE{G}{h}$---one estimate for every hyperparameter configuration $h$---and we select the estimate that is the largest in order to produce $\max_h \CE{G}{h}$.
%However, any hyperparameter for which we overestimated the average performance will have an unfair advantage during this maximization process.

To gain a bit more intuition, consider the following example comparing two hyperparameter configurations $h_1, h_2$.
Imagine that $\CE{G}{h_1} = \CE{G}{h_2}$, but when we estimate their average performance using samples, $\bar{G}_{h_1}$ is an overestimate of $\CE{G}{h_1}$.
When we maximize over our estimates $\bar{G}_{h_i}$, we will be reporting an overestimate for $\max_h \CE{G}{h}$ meaning we are overstating the performance of our algorithm. This is similar to maximization bias we see in Q-learning!
Now imagine instead of two configurations that are identical, we instead have 100 identical configurations, or 1000.
The probability that we overestimate at least one configuration increases as we increase the number of configurations estimated; thus increasing the reported performance for the algorithm, $\max_h \CE{G}{h}$.
That is, without changing the properties of the investigated algorithm itself, we can report increasingly higher performance by increasing the number of hyperparameter configurations we investigate!
%ADAMC: removed a bunch of this because I didnt think it was needed
%In fact, the above example occurs frequently even in simple reinforcement learning experiments.
%If we run 1000 experiments of DQN on the Mountain Car environment and sweep stepsizes, target network refresh rates, and replay buffer sizes for every experiment, then approximately 96\% of these experiments will over report the average performance for the best hyperparameter configuration among those tested.
%As a result, we no longer have an accurate assessment of the performance of DQN on Mountain Car.
If this was an algorithm that we are proposing, then we would be doing our readers a disservice by overstating the benefits of the algorithm.
If this was a baseline algorithm, we would be doing ourselves a disservice by setting too high of standards and potentially filtering out useful ideas.

\paragraph{Issues with the typical two-stage approach.}
\label{2-stage}
One common approach to counteract the effects of maximization bias is to use a two-stage methodology.
In the first stage, the researcher performs an extensive hyperparameter sweep in order to select the maximizing hyperparameters.
In the second stage, the researcher then evaluates the best hyperparameter configuration with a new set of random seeds---typically far more random seeds than originally used for the selection process.
Finally, we report the average performance for the second stage only.
This approach provides an unbiased, confident estimate of performance for that chosen hyperparameter configuration, because we are able to use more runs for that single hyperparameter configuration.

However, there are two key drawbacks to this approach. As pointed out by \citet{jordan2020evaluating},
this approach is wasteful of compute, because many samples of performance are thrown out during the hyperparameter selection stage.
Additionally, this two-stage approach ignores uncertainty during hyperparameter selection.
Say, for example, that we use 10 runs for every hyperparameter configuration in stage one.
%The performance estimates for each configuration will likely have a large amount of error; however, hopefully the estimates are accurate enough to approximately order the configurations from best to worst.
Due to maximization bias, however, we may unintentionally select a suboptimal hyperparameter configuration as being best.
Then in the second stage, we will use many runs---say 100---to evaluate this hyperparameter configuration.
Because we used so many runs in this second stage, our confidence intervals about the mean performance will be tight---we will be reasonably certain we have accurately captured the mean performance.
This certainty, however, is unwarranted.
While we may have accurately estimated the mean for the selected hyperparameter configuration, we likely have not accurately estimated the mean for the \emph{best} hyperparameter configuration.
As a result, this two-stage approach is likely to underestimate the maximum performance and is overconfident in its estimate.

Regardless, the two-stage approach is a reasonable place to start and is much better than one-stage tuning or ``tuning by hand"---both common in the literature.

\paragraph{A new approach: bootstrapped two-stage tuning.}
%Consider the typical approach to investigating hyperparameter sensitivities; sweeping over all combinations of hyperparameter configurations.
%For every configuration, we produce an estimate of the average performance.
%We could simply select the configuration which has the highest estimated average performance and report this performance alongside a confidence interval, however as we discovered previously, this will systematically over-estimate the performance of our algorithm.

One way to overcome maximization bias is to repeat the entire experimental procedure multiple times; wrap the two-stage process in an outer loop. After running the sweep over hyperparameters the first time, we repeated this procedure a second time using a new set of random seeds.
We will almost certainly obtain a new estimate for maximal performance and likely obtain a different best hyperparameter configuration that gives that maximal performance.
We now have two different estimates of maximal performance, giving us a sense of spread---how much does this maximizing performance change if we use different data to estimate it?
Naturally, we can repeat this process many times---say 100---and report the mean and spread of the results.

Unfortunately, this simple procedure has a major downside: it is incredibly expensive.
For every hyperparameter configuration (say $H$ configurations), we require $N$ runs and we repeat the sweep-then-maximize procedure $M$ times.
In total, we run our learning algorithm $H \times M \times N$ times in order to report an estimate of tuned performance.

Fortunately, we can rely on the principles of bootstrapping and resampling in order to make this procedure much cheaper.
For every hyperparameter configuration, we collect $N$ runs of our algorithm.
Then, for each configuration, we resample $N$ of those $N$ runs (with replacement) and compute a sample average.
We select the maximizing sample average as a single estimate of tuned performance.
Then we repeat the above procedure many times, resampling from the same $H \times N$ dataset each time.
Because we are capturing the variance across all hyperparameters, we do not need as many runs \emph{per} hyperparameter.
As a result, we can select a smaller $N$ than we would typically require for sensitivity analysis; say $N=10$ instead of $N=30$.

\paragraph{Picking hyperparameter sets fairly.}
We now have a mechanism for measuring idealized performance, but still want to avoid the pitfall where we allow one algorithm to have many more hyperparameter settings than another. We want our results to reflect the utility of our algorithm, rather than reflecting performance when fitting hyperparameters to a set of environments. Unfortunately, there is no an explicit, precise procedure here to obtain a perfectly fair experiment. The only way to avoid tricking ourselves is to attempt to be as fair as possible and to sincerely make choices that are justifiable.

At a minimum, you should ensure all algorithms have the same number of hyperparameter settings that are tested. If one algorithm has two hyperparameters, $\alpha$ and $\beta$, and another only has $\eta$, then you have to sweep $n$ values for $\eta$ and $n$ values for the cross-product of $(\alpha, \beta)$. For example, you might test $\eta \in 2^i$ for $i \in \{-6,-5,-4,-3,-2,-1\}$ and test $\alpha \in 2^i$ for $i \in \{-5,-3,-2\}$ and $\beta \in \{0.1, 0.5\}$, so that they both have 6 hyperparameter settings.

If you are using a hyperparameter optimization algorithm, then you can optimization algorithm to ensure fairness. These optimization algorithms typically attempt to estimate the function mapping hyperparameter values to performance for a given agent-environment pair (e.g., using bayesian hyperparameter optimization in Optuna). On each iteration, the optimization algorithm samples a new hyperparameter configuration based on the learned function so far (selecting the one believed to be best, with some exploration) and then runs an reinforcement learning experiment---for some number of steps and repeating the experiment with different random seeds---to obtain a performance estimate.
When using such a tool, we can ensure fairness by ensuring the optimization algorithm is run for a fixed number of iterations and the same number of steps and seeds are used in evaluation. Automatic hyperparameter selection is a big topic and there are several great papers you can start with if you want to learn more ~\citep{he2021automl,parker2022automated,eimer2023hyperparameters}. We discuss fair comparisons of multiple agents in much more depth in Section \ref{sec_comparing}.

\subsection{Evaluating Algorithms for Deployment}\label{sec_hyperdeploy}
This final setting evaluates the performance of a reinforcement learning algorithm under some ``realistic'' constraints, typically modelling a specific deployment scenario.
For this setting, we need an \emph{algorithmic} approach to set hyperparameters, or we need hyperparameter-free algorithms. It is not possible to test many hyperparameters in most deployment scenarios.\footnote{
  The one exception is when the ultimate goal is to solve a simulated environment.
  This is beyond the scope of this document, but addressed elsewhere ~\citep{parker2022automated,eimer2023hyperparameters}.
  It is important to note that for simulated environments, even though we can use hyperparameter optimization (or sweeps), we then have to account for that as part of the algorithm and as part of the computational cost.
}
For example, if a reinforcement learning algorithm is being used to optimize data center cooling, it can be unrealistic to test even a handful of hyperparameters on the system. Instead, we want to deploy a fully-specified algorithm.

One issue in reinforcement learning is that we do not have a general purpose algorithm for hyperparameter selection, unlike supervised learning. In supervised learning, we can use internal cross-validation to select hyperparameters: the best hyperparameters are selected by separating the data into training and validation, and using validation performance as a measure of generalization performance under those hyperparameters. We have no such equivalent procedure for reinforcement learning. We discuss why this is the case in Appendix \ref{app_cv}, and propose potential options for developing such algorithms for reinforcement learning.

This final setting is an open question in the reinforcement learning community.

A common approach today is to use the default hyperparameters specified in released code-bases.
Using these defaults is not unreasonable, if the goal is to compare two systems. In an empirical study, you may want to understand the performance of two code-bases across a variety of different environments. This experiment is not about comparing the algorithms underlying those systems, but rather the systems themselves. It could help a practitioner decide which of the available code-bases might be more suitable for their application. Of course, the default hyperparameters in the code-base were likely set on a small set of simulation environments; so we should be cautious about how well the system will perform in deployment.

Another strategy has been to tune hyperparameters on a subset of environments, and then fix them for a larger set of environments. This practice was used in the Atari suite for example, where it was suggested to use five of the 57 games for hyperparameter tuning \citep{bellemare2013arcade}.This procedure could mimic a deployment scenario, where you have several simulated environments related to your real-world environment, on which hyperparameters can be tuned. However, as yet there is little understanding of how one might pick such tuning environments. %, nor issues with the disconnect between the tuning set and the deployment set of environments.

Finally, a simple and relatively under-investigated approach is to learn a simulator of the real system and use the resultant simulator to search for or optimize the hyperparameters~\citep{wang2022no}. The advantage of this approach is the number of hyperparameters tested, in a sweep or bayesian hyperparameter optimization proceedure, is in no way limited by the deployment environment. The challenge of course is that the quality of the simulator is related to the deployment data available and, generally, learning accurate models that are useful under long rollouts remains a largely open challenge.
\newpage
%Finally, some work has focused on picking one set of hyperparameters across a range of environments, rather than selecting different hyperparameters per-environment. This differs from the subset approach discussed above, because we are trying to tune across all deployment environments. Such a procedure can help identify suitable default parameters, or identify how to change the algorithm to remove or automatically adapt hyperparameters. It is a step towards obtaining a fully-specified algorithm, rather than a general purpose approach to select hyperparameters for an algorithm that is not fully-specified. Nonetheless, this might be the right direction for handling hyperparameters in reinforcement learning: obtaining parameter-free algorithms (fully-specified algorithms) or those for which hyperparameters are easy to tune based on the knowledge of a given environment.

\begin{summary}[Key insights: dealing with hyperparameters]{}

%We summarize the key pieces of actionable advice about hyperparameters here.
\begin{enumerate}[leftmargin=7pt]
\item Unthoughtful treatment of hyperparameters is one of the leading causes of bias in empirical reinforcement learning. The first step is acknowledging this bias.
\item When designing an experiment, make a plan to select hyperparameters in a way that matches the goals of the experiment, rather than as a frustrating nuisance.
\item Hyperparameter sweeps are not a general-purpose method for finding hyperparameters; instead, we use sweeps in experiments to understand our algorithms.
\item For sensitivity plots, select ranges of hyperparameters mindfully. Ensure the range is wide enough; if you find optimal parameters lie on the boundary, expand the range.
\item Visualizing hyperparameter sensitivity is itself a difficult problem; be innovative about how to communicate this to the reader. Current options include two-dimensional parameter sensitivity curves over one hyperparameter and violin plots of all hyperparameters.
\item In general, it is better to avoid reporting performance for the best hyperparameters chosen in an environment. There are too many pitfalls, and you risk tricking yourself about the quality of your algorithm.
\item Default hyperparameter settings are not necessarily appropriate and certainly not always fair. See more in the next section about the pitfalls of untuned baselines.
% MARTHAC: I removed this since actually we just dont have the best advice for this yet
%\item To design experiments that reflect performance in deployment, try to consider how hyperparameters may actually be selected in practice. This might involve tuning across environments to avoid overtuning, or tuning on a subset of environment to mimic translating learning in algorithm design to deployment.
\end{enumerate}
\end{summary}

\biblio

\section{Comparing the performance of multiple algorithms}\label{sec_comparing}
Most---if not all---the things we worry about when investigating a single agent are relevant when investigating more than one.
However, the concerns become more serious as we are often making a value judgement on the ranking or relations between multiple agents.
The claims are inherently more nuanced and thus the standard of evidence and rigour required goes up a notch.

There are many reasons we ultimately compare multiple algorithms.
The most common goal is to provide a performance ranking over multiple related algorithms in a given environment (or suite of environments).
Implicitly, we are making the claim: ``if your environment is similar to this environment, here is the algorithm you should use''.
Supporting such claims well is rife with difficulties and we will dive into the details in Section~\ref{sec_environment_selection}.

An alternative reason to compare algorithms is to show that one algorithm (the baseline) suffers from some problem, but the newly proposed algorithm does not.
A prototypical example is the introduction of Gradient TD algorithms \citep{sutton2009fast}.
The comparison starts by convincingly illustrating that the baseline algorithm (TD in this example) suffers from some problem by crafting a very specific and simple counterexample.
Then we introduce a novel algorithm or modification to the baseline and show that this modification does not suffer the same problem.

In this section, we will explore challenges that arise for both types of comparison and discuss strategies for drawing reliable and robust conclusions.
%Naturally, it is impossible to capture all possible future experiments, however, much of the following discussion generalizes well to many possible experimental procedures.
We will assume the reader has well understood the preceding sections on summarizing the performance of a single algorithm (Section~\ref{sec_single_more}) and understanding the impact of hyperparameters on an algorithm (Section~\ref{sec_hypers}) as these concerns become even more exaggerated in the multiple algorithm case.

\subsection{Designer's curse: with great knowledge comes great responsibility}\label{sec_untuned}

Evaluating your own algorithm is rife with bias.
You know the ins and outs of your algorithm better than anyone on the planet. You have spent months working with different versions of your algorithm, tuning them for performance, finding environments where your algorithm shines, and discarding ones where your algorithm does not. The baselines you eventually compare against likely have not received the same attention. Worse, you will not have the same detailed understanding of those baselines, nor knowledge of how to tune them well.

There is nothing we can do about designer bias %(in fact, it is a nice side effect of good algorithmic research)
, but there are steps we can take to reduce your experimenter bias.
You can counteract some of the bias, however, by spending extra effort to get the baselines working well. It is better to risk giving the baselines a small advantage to mitigate over-claiming.

Let us revisit the example of an experiment where we propose an algorithm to resolve some specific failure case.
Implicitly, this experiment is actually making two independent claims.
The first claim is that the baseline algorithm experiences failure.
As we saw in Section~\ref{sec_hypers}, it is exceptionally challenging to show that an algorithm has a consistent failure case.
We can always ask: \emph{would this happen for a different choice of hyperparameters?}

The second claim is that our proposed algorithm solves the failure case.
Illustrating this second case can be much easier; we just need to find a single hyperparameter setting that works!
As a result, a truly fair comparison requires spending a significant amount of time understanding the baseline algorithm while requiring far less time (experimentally) understanding the proposed algorithm.
If we had not spent the time with the baseline, we could unintentionally build a strawman argument; one which does little good for the scientific understanding of learning algorithms.

When we are benchmarking our algorithm, one of the easiest ways to protect against designer bias is to choose the right environments.
We can let the authors of the baseline algorithm do the work here.
If you are comparing your new algorithm to \emph{Soft Actor-Critic} (SAC) \citep{haarnoja2018soft}, then use an environment where prior work has shown SAC does well.
It is reasonable to assume that the authors of SAC worked hard to tune their own algorithm---that is designer bias working for you!
It is important that you pair the environment choice with fully specified algorithms because our algorithms are not yet general. SAC with hyperparameters tuned for HalfCheetah, for example, is likely to not work well in Mountain Car. It is okay to try a baseline algorithm on a new environment, but then you will likely have to expend significant energy tuning it (perhaps changing the network architecture, optimizer, etc.) and you really cannot be sure you will do a good job.

\newpage
\begin{myexample}[Untuned baseline.]\label{example_untuned}
  \begin{wrapfigure}[9]{l}{0.3\textwidth}
    \centering
    \includegraphics[width=0.9\textwidth]{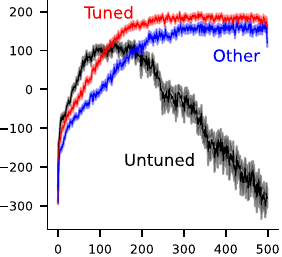}
  \end{wrapfigure}
  The inset figure shows one example where an untuned agent is applied to a new environment, in this case Lunar Lander. Imagine we have introduced a new algorithm called DeepQ---it is a simple action value learning agent with a Q-learning update and neural network function approximation (the blue line). Now imagine we grabbed a freely available DQN implementation (the black line) with a default set of hyperparameters tuned for some other environment. Comparing the blue and black lines might lead one to think DeepQ is much better than DQN on this task, but we are misleading ourselves. The red line represents DQN after retuning its hyperparameters for Lunar Lander. Much improved performance and ego checked!
  %We will need to measure alternative metrics or rethink our research question because on this new evidence we must reject the hypothesis that DeepQ is better than DQN on Lunar Lander.
  The moral of the story is simple: \textbf{beware of untuned baselines.}
\end{myexample}

\subsection{The utility of calibration baselines: what does that line even mean anyway?}
Sometimes comparing the performance of two algorithms is not enough because we lack context to understand the results.
Returning to the Lundar Lander experiment above we might naturally wonder: are any of these algorithms doing well? How would a non-learning policy like random action-selection fair?
What about a learning agent with a much simpler function approximator like SARSA($\lambda$) with tile coding?
Anecdotally, years ago when one of the authors submitted their first paper on using reinforcement learning to play hearts, pesky Reviewer \#2 said roughly: ``sure your agent is good compared to human players and the best search agent, but I don't have any basis to understand the performance. You should include results playing against a random agent''. Reviewer \#2 was right. We included the baseline, which ultimately showed our agent and the other baselines were very strong indeed, and it made our paper better. You should always ask yourself if such baselines---not just SOTA algorithms---could make your results easier for the non-expert to interpret.

Picking the right calibration baseline depends on the research question, but generally there is a lot of flexibility here. It is often good to think of both randomized and high-performance or even oracle baselines.
Oracle baselines often have access to side information that your learning agent does not, setting an unobtainable but interesting performance bar. Alternatively, an oracle baseline might just operate under different constraints than your new algorithm. For example, we often compare linear complexity off-policy TD algorithms against  least squares TD (LSTD). LSTD is a quadratic algorithm that performs more compute per step than a linear algorithm and thus naturally we expect it to set a high performance bar. If your new linear off-policy TD algorithm approaches the performance of LSTD on several environments, you can be more confident your algorithm is learning efficiently and that your algorithm's improvement over other linear baselines is relevant.

While these calibrations provide useful context for interpreting results, we must be cautious not to become over-reliant.
For example, the Atari suite uses random performance and human-level performance as calibration baselines \citep{mnih2013playing}.
However, humans have rather different constraints than our agents; requiring a screen to project light to our eyes, having latency between our brains and fingers, the often imperfect translation from muscle activation to controller input, and finally latency from controller to gaming system.
Achieving human-level performance is certainly informative of learning capacity and progress; exceeding human-level performance by many orders of magnitude, however, may be a result of fewer constraints.
As we move increasingly far from our calibration baselines, we must re-evaluate their utility.

\subsection{Ranking algorithms (maybe don't)}

A common goal is to show that your new algorithm is better than some prior work.
However, providing supporting evidence towards this claim can be exceptionally difficult. In fact, in all but the most rare cases, we might go so far as to say that providing such evidence is entirely infeasible!
The space of problem settings where any given algorithm may be deployed is large and wholly unspecified; showing that any algorithm generally outperforms another across this space would be impossible without first defining the space.
Even once this massive problem-space is well defined, gathering sufficient evidence to cover the space is likely intractable.

Instead, we tend to produce a ranking of algorithms on a small set of environments and hope our claims generalize. Let us discuss a bit more why this might be problematic.
Even if we could show improved performance for a particular benchmark, this is likely not broadly informative.
It is like trying to generalize from a biased distribution (say performance in Atari) with a few samples to make predictions about performance across the whole set (performance in all environments).

Instead, we argue that insights about \emph{why} an algorithm behaves differently tend to generalize. {\em Scientific testing} is one name for designing experiments to gain insight on \emph{why} questions and is nicely discussed in \citet{jordan2024position}. Scientific testing is often performed using synthetic environments, like counterexample MDPs or environments with carefully crafted properties such as Riverswim \citep{strehl2008analysis}. These synthetic environments allow us to reason about what situations \emph{will} cause an algorithm to fail.
It is up to domain experts, then, to identify if these failure modes apply to their problem setting.

\begin{myexample}[The deadly triad.]
  One such insight that has aged well is the \emph{deadly triad} \citep{sutton2018reinforcement}.
  The deadly triad specifies three conditions where TD-based algorithms often fail. Specifically, when the algorithm uses (1) bootstrapped estimates, (2) off-policy sampling, and (3) function approximation.
  Given this set of conditions, then, it is not hard to propose an environment where TD algorithms fail; many have been proposed in the literature \citep{baird1995residual,kolter2011fixed,tsitsiklis1997analysis}.
  These conditions have been further refined over the years to include properties of the environment \citep{kolter2011fixed,patterson2022generalized} and properties of the function approximator \citep{ghosh2020representations} allowing an even more crisp understanding of when algorithms might fail.
  %These insights have later been shown to be robust across many learning problems, architectures, and fully engineered learning systems \citep{vanhasselt2018deep,obando-ceron2021revisiting,patterson2022generalized}.
\end{myexample}

Another challenge that arises is that of fairness.
When comparing the performance of multiple algorithms, many design choices must be made: setting hyperparameters, picking which environments to use, how long to run each experiment, and so on.
Sometimes these choices have non-obvious and indirect impacts on the performance of each tested algorithm, often leading to latent unfairness in the experimental design.
A noteworthy example is the experiment length used for the Atari benchmark.
Shortening the length of experiments run on Atari dramatically changes the ranking of algorithms dramatically ~\citep{machado2018revisiting,agarwal2021deep,obando2024consistency}. It remains unclear if Atari 100k simply favours agents that learn much more aggressively (e.g., larger learning-rate parameters) without risk of divergence, or if it captures something different and interesting compared with the usual 200 million step setup. %leading one to wonder which ranking to consider when applying any of these algorithms to your problem of interest.

% --> General strategies for ensuring fairness
Sometimes it is not obvious how to maintain fairness across different algorithms.
%In these cases, a common strategy is to make comparisons to multiple versions of an algorithm.
%at multiple levels along the challenging axis.
For instance, if two algorithms have wildly different architectures, it can be challenging to ensure representation capacity is comparable.
Instead, an empiricist might fix the representation capacity of one algorithm, while testing the other algorithms with increasingly large representations.
One could then report the performance with different representation sizes and thus give a more complete picture of the performance difference as representation capacity changes.

Unfortunately, it can be expensive to exhaustively test an algorithm for multiple settings of a confounding variable---in fact, this is effectively a form of sensitivity analysis akin to Section~\ref{sec_hypers}. When designing fair comparisons, we often come back to focus on understanding our algorithms, rather than ranking them. We discuss this further in Section \ref{sec_environments}, where we discuss selecting and designing environments for experiments.

If this cost is prohibitive, an alternative option would be to explicitly run an unfair experiment, providing an explicit disadvantage to our own proposal algorithm.
Often, our proposed algorithm already has multiple sources of implicit unfairness: we spend more time on the implementation, we better understand the algorithm and when it may fail or succeed, we spend more time tuning the algorithm to our problem setting.
When we provide an explicit source of unfairness---such as a reduced representation capacity---and our algorithm still outperforms competitors, we show a lower bound on the potential improvement provided by our algorithm, implying that an even greater degree of improvement would be observed under even more fair conditions.
Naturally, we should be careful not to overclaim here; we cannot know to what extent our algorithm improves performance under more fair conditions.

\subsection{Statistically significant comparisons for two fully-specified algorithms}\label{sec_twoalgs}

Imagine that we want to compare Algorithm A and Algorithm B, in Mountain Car, in terms of the online episodic return, in expectation across many runs. We may want to say that A is better than B, with high confidence, on Mountain Car.\footnote{Presumably we are verifying that A resolves an issue with B, that manifests on Mountain Car, rather than attempting to outperform B on Mountain Car.} One of the simplest strategies for comparing the two is to compute confidence intervals, which we discussed for a fully-specified algorithm in Section \ref{sec_ci}. If the two intervals do not overlap, and the mean value for A is above B, then we can conclude that A is statistically significantly better than B.

However, this is a low powered test. In other words, depending on the number of runs, the confidence intervals may overlap but a more powerful test, like the paired t-test, might have allowed us to conclude that A is statistically significantly better than B. The primary reason is that pairing allows us to account for sources of variation due to the environment or initialization. For example, for one random seed, the agent may start in a difficult start state, impacting all future learning. If we compare the two agents for that seed, then we may find that both performed poorly for that seed, but that the relative ranking remained the same. The small modification to get the paired t-test is simply to look at \emph{differences} in performance, rather than the performance itself.

\begin{figure}
  \centering
  \begin{subfigure}{.49\textwidth}
    \includegraphics[width=0.85\linewidth]{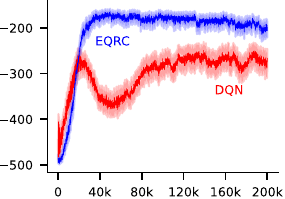}
    \caption{\label{fig:mc-ci-curves}
      Individual performance of DQN and EQRC.
    }
  \end{subfigure}
  \hfill
  \begin{subfigure}{.49\textwidth}
    \includegraphics[width=0.85\linewidth]{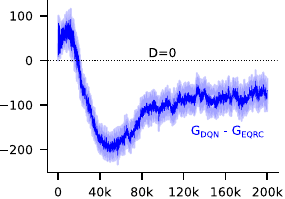}
    \caption{\label{fig:mc-diff-curve}
      Performance difference $D = G_\text{DQN} - G_\text{EQRC}$.
    }
  \end{subfigure}
      \caption{\label{fig:mc-diff-example} {\bf Differences can make a difference:} in \textbf{(a)}, we see the performance of two algorithms, DQN and EQRC \citep{ghiassian2020gradient}, on the Mountain Car environment.
      Shaded regions show individual confidence intervals for each algorithm. In \textbf{(b)}, we show the performance difference $D = G_\text{DQN} - G_\text{EQRC}$.
      When $D > 0$, then DQN has better performance.
      When $D < 0$, then EQRC has better performance.
      And when the shaded region \textbf{does not} include the horizontal line at $D = 0$, then the reported difference is statistically significant.
      Looking at the learning curves on the left (a), it looks like the performance of the two algorithms is nearly indistinguishable in early learning. Looking at difference curves on the right (b), we get a clearer picture.}
\end{figure}

We can leverage the same idea to visualize confidence intervals for learning curves.
Overloading terminology, let the performance of one algorithm be random variable $A$ and the performance of the baseline be random variable $B$. Then we would report an interval around $D = A - B$ as opposed to reporting two intervals $(l_A, u_A)$ and $(l_B, u_B)$, as shown in Figure \ref{fig:mc-diff-example}.
Whenever the lower bound of the interval is greater than zero, our proposed algorithm outperforms the baseline with our stated level of confidence.
Notably, this has the additional advantage of removing one line and shaded region in learning curve plots---at the cost of a slightly obfuscated visualization of performance. In the end, both plots may be desirable.

It may actually be preferable to use this confidence interval on the difference rather than a  hypothesis test because it provides more information. In particular, it highlights the magnitude of the difference between algorithms, called the \emph{effect size}, not just that they are different.
Most hypothesis tests are designed to answer the question: is the average performance of algorithm A better than the average performance of algorithm B?
If there truly is a difference in performance between these algorithms, then with enough samples we will eventually be able to reject the null hypothesis and provide evidence that A is better than B at the $p=0.05$ level.
However, this ignores a very important nuance: the difference in performance may be negligibly small.
A common solution to this problem is to include an appropriate measure of the effect size for the chosen hypothesis test, for instance for a t-test one might include Cohen's $d$ to measure the effect size. However, such effect size tests may not be available in all situations and can often be difficult to interpret.

Another common criticism of hypothesis tests are that their results are easily misinterpreted; in fact, this has led to some scientific publications banning the use of p-values \citep{wasserstein2016asa}.
A common misinterpretation of the p-value is that $p$ represents the error rate of the test.
Rather, the $p$ value says: \emph{if the null hypothesis were true, then the probability of seeing results at least this extreme due simply to randomness is $p$}.
However, naturally, we do not know whether the null hypothesis is true---this is why we are performing the test---and so we cannot know the true error rate from this alone.
Recent statistical simulations suggest that, for sensible assumptions, the true error rate for a p-value of $p=0.05$ may be as high as $26\%$ \citep{sellke2001calibration,colquhoun2017reproducibility}.
As the p-value decreases and the effect size increases, this error rate quickly drops.

The principle behind pairwise comparisons is also about controlling randomness whenever we can.
The ability to replicate and reproduce results is core to scientific investigation.
Computer simulations have a particular advantage of being nearly perfectly replicable.
To do so, however, we must carefully control sources of variation: this means controlling the pseudorandom number generators responsible for simulating randomness in our environments and agents.
A common approach is to set a global random seed for the entire simulation, impacting the random state of all randomized components simultaneously.
However, it can often be beneficial to retain independent random states for each randomized component of the experiment. We give an example below about how this could be useful when evaluating the addition of auxiliary losses, measuring a property called the \emph{stable rank}.\footnote{The stable rank is the ratio between the Frobenius norm and the spectral norm of the weights. It was introduced as a measurement of capacity, for instance \citet{arora2018stronger}.}

\begin{myexample}[Separating the random seed for the agent and environment.]\label{example_repeated}
Imagine you want to understand the impact of adding an auxiliary loss to your agent.
  Specifically, your hypothesis is that the stable rank of the agent's representation will decay less quickly when you use an additional neural network head to predict the next state.
  To test this hypothesis you start with a single environment---say Puddle World---select a learning algorithm with all hyperparameters specified except for the neural network architecture, then create two fully-specified algorithms by defining a neural network architecture with the auxiliary head, and another without.
  Finally, you run each algorithm 30 times, measuring the stable rank of the neural network layers periodically through the run for each individual agent.

  Consider the code-design where you set a single global random seed for the entire simulation and you run the first $\text{seed}=0$ for both architectures.
  Because both neural networks have different sizes and both are initialized randomly, each agent will call the random number generator a different number of times---at the end of agent initialization, both random number generators are in different states.
  Then your code initializes the Puddle World state by randomly selecting a starting coordinate.
  However, because both RNGs are in different states, both investigated agents likewise start in different states.
  Because the starting state in Puddle World considerably changes the sequence of observed rewards, the agents for $\text{seed}=0$ receive wildly different data.

  On average over several different agents for each experimental condition, the effects of differing observation sequences will wash out and we will be able to detect if our auxiliary loss plays a role on stable rank.
  However, if we had used individual random states for both the agent initialization and the environment initialization, the two $\text{seed}=0$ agents would have observed the exact same sequence of observations and would share some degree of variation due to these observations.
  Taking advantage of this pairing structure, we can detect differences in stable rank with far fewer agents by cancelling out nuisance sources of variation.
  This experimental design is called \textbf{repeated measures}.
\end{myexample}

\subsection{Statistically significant comparisons for multiple fully-specified algorithms}

There are times we want to compare multiple agents. For example, there may be a baseline, an existing algorithm and your modification of that algorithm. A sensible choice is still to choose an algorithm as a comparator, and plot differences to that algorithm for the rest. This approach mitigates some of the variance we might see due to the particular configurations in a run. We can report confidence intervals for these difference curves, as above.

However, we need to consider issues with \emph{multiple comparisons}. Imagine that we compare five different algorithms on a plot, all with performance differences to some baseline. We may mentally compare the five different confidence intervals, or run pairwise comparisons using hypothesis tests. If we treat each comparison independently, then the failure probability $\delta$ accumulates. To account for these multiple comparisons, the simplest solution is the Bonferroni correction \citep{dunn1961multiple}, that uses $\delta/5$ to compute these accumulating probabilities to allow for a confidence of $1-\delta$. Naturally, this requires even more runs (and more compute) to ensure statistical significance.

One of the simplest solutions to this issue is to narrow the scope of our experiment, and run fewer algorithms. We can limit the scope of baselines to those that have similar design philosophies or characteristics to our own proposed algorithm.
% MARTHA: I don't think we should say this
%(showing reviewers at times request more baselines with little thought).
For example, perhaps we propose a model-free policy-gradient algorithm and so choose to only compare to other model-free policy-gradient baselines. This restriction narrows the question to algorithmic improvements within the same class of algorithms, rather than also trying to show improvements for policy gradient algorithms over an entirely alternative class like model-based Q-learning algorithms. It is actually reasonable to separate this for clarity reasons, not just computational reasons.

% MARTHAC: I am not sure we should include this.
If, on the other hand, your empirical goal is to compare a large set of algorithms, then the answer is simply that you need to do more runs. It can be useful to investigate if policy gradient algorithms have advantages over an alternative class like model-based Q-learning algorithms. When answering such a broad question, naturally it requires more compute. This further highlights why it is useful to separate these concerns. It is worthwhile to have empirical studies and benchmark papers that attempt to run these more comprehensive experiments. This is a mountain of work! It is perhaps infeasible and not entirely helpful to also include such work within a paper focused on a new algorithm. Simply put, ever paper need not include a set of leaderboard-style results.

Another common approach is to isolate the precise modification made in the proposed algorithm.
Perhaps the proposed algorithm introduces a new sampling strategy from replay buffers.
Instead of comparing against a suite of different learning rules, we can instead make a small number of pairwise comparison between similar algorithms.
For instance, we can endow DQN and A3C with our proposed replay buffer and perform two pairwise comparisons against the original DQN and A3C respectively.
We do not care if DQN with the novel replay buffer outperforms A3C with the old replay buffer, as many design variables change between these populations.
By eliminating several such pairwise comparisons, we can significantly decrease the probability of false negatives and number of samples to detect differences.
%which in turn decreases the number of samples required to detect changes in performance due to the proposed modifications.

\begin{summary}[Key insights: comparing the performance of multiple algorithms]{}

%We summarize the key pieces of actionable advice about hyperparameters here.
\begin{enumerate}[leftmargin=7pt]
\item Remember, the seed is NOT a tuneable hyperparameter.
\item Be aware of your designer bias. Ultimately, you want to avoid misleading yourself, as well as the rest of the community.
\item Consider the use of baselines (e.g., random, oracle) to contextualize performance.
\item The purpose of comparing to an algorithm is primarily to show that your new approach resolves the issues that you claim it does. It is difficult to claim that you have a generally better algorithm only using performance across several environments. Design experiments to provide insights rather than state-of-the-art claims.
\item We can leverage paired comparisons between algorithms, to reduce variability and make statistically sound conclusions with fewer runs.
\item Statistical intervals---confidence intervals and tolerance intervals---provide a more interpretable alternative to hypothesis tests, when comparing two algorithms.
By reporting intervals, we provide both a sense of effect size and an estimated error rate.
\item Consider setting the random seed for the environment and agent separately. This \emph{repeated measures} empirical design allows us to control for stochasticity in the agent (e.g., neural network initialization) separately from environment stochasticity (e.g., start state), ensuring pairwise comparisons share more similarities.
\item Comparing more than two algorithms requires more care, and we can run into \emph{multiple comparisons} issues. Err on the side of designing experiments where a more focused questions is asked ({\em i.e., scientific testing}) comparing a smaller number of algorithms.
\end{enumerate}
\end{summary}

\biblio

%\subfile{parts/quick_start.tex}

\section{Selecting environments}\label{sec_environments}

The selection of environments is a critical part of the setup of your experiment. In other sciences, experiments are about the natural world and are naturally constrained.
The natural world provides a rich suite of problems for us as well, such as robotics, chat bots, controlling power plants, trading, and more.
These real-world problems are often where we want to eventually deploy our algorithms; simulation is where we can prototype. In simulation, it is difficult to maintain the richness of the real-world. And worse, we may inadvertently design environments that are actually impossible or too easy. %We have a lot of flexibility to understand our agents in simulation, and with that flexibility comes hard questions. We may know what we want to test about our algorithms, but i might not be easy to how do we either find or design an environment to do so?

When developing an algorithm, the first step is usually to hypothesize a very simple environment and experiment where you are highly confident in the outcome. For example, to test an agent's ability to remember, we might design a small gridworld where the agent must remember that pressing a button unlocks a reward. This first step provides a foundation for more complex experiments. If there are surprising results in this first step---as there very often are---then it can be much more carefully understood before moving to a setting where it can be harder to isolate the issue. It is also a step where a surprising result can make you rethink the algorithm itself, thus providing conceptual clarity.
We discuss such \emph{diagnostic} environments in Section \ref{sec_diagnostic}.

After that, the next step is to evaluate the algorithm(s) on existing benchmark environments. We discuss the role of using benchmark environments further in Section \ref{sec_environment_selection}. We conclude the section with a brief discussion about difficulties with creating new environments and about the potential utility of aggregating environments.
% in Section \ref{sec_aggregating}.

\subsection{Diagnostic environments}\label{sec_diagnostic}

The first step in designing a diagnostic environment is to isolate the key issue your algorithm is designed to selve. If you can identify that issue, then you can design an environment that is defined by that issue or where the issue is exaggerated enough to find an effect---that is exactly what a {\em diagnostic environment} is.

A classical diagnostic environment is Baird's counterexample \citep{baird1995residual}.
This 7-state environment was designed to show that temporal difference (TD) learning divergences under off-policy sampling.
This diagnostic environment may seem like a counterexample, rather than an environment, however it is an ideal example of a diagnostic environment because it is precisely one where the experimenter should be quite sure of the outcome, divergence.
Surprising outcomes have since come out of this environment, for example, we have found that incorporating certain adaptive stepsize algorithms or periodically fixed targets (e.g. target networks) into TD seems to resolve this counterexample, though there is no theory that suggests these strategies should help.

These diagnostic environments can have lasting impact on algorithm development. In some cases, Baird's little MDP has highlighted issues with algorithms that claim convergence guarantees; for example Emphatic TD should converge on Baird's counterexample in theory, but in practice the variance is so high that convergence is very poor and requires careful tuning \citep{sutton2016emphatic,mahmood2017incremental}.

Diagnostic environments are chosen based on the hypothesis you wish to test, which means that an existing diagnostic environment may already have been designed to isolate that property.
For example, you might hypothesize that an algorithm is prone to settling on a suboptimal policy because it does not explore enough, so you might use Riverswim \citep{szita2008many}, which was designed specifically to test this.

\subsection{Benchmark environments and challenge problems}\label{sec_environment_selection}

Benchmark environments provide a useful tool to assess if there are issues with a new algorithm---a sanity check. The word \emph{benchmark} means a point of reference. The possible performance on these environments is well understood, due to previous results, making it easier to understand if you are doing better or worse than this reference. If you are doing notably worse than a known reasonable solution, then this could indicate an issue with your algorithm that should be addressed.

For this reason, benchmark environments can remain useful for many years.
Classic environments that are still commonly used include Mountain Car \citep{moore1990efficient}, Cartpole \citep{sutton2018reinforcement}, Puddle World \citep{sutton2018reinforcement} and Acrobot \citep{sutton1996generalization}. These simple environments play a useful role because we understand very well how to get good performance in these environments, and so we can easily see when there are issues in new algorithms. For example, a SARSA($\lambda$) agent with tile-coding can find a good policy in Mountain Car in a dozen or so episodes. DQN---using replay and target networks---learns much more slowly, settles on a worse policy, exhibits more instability, and is sensitive to the target network refresh rate \citep{hernandez-garcia2019understanding,kim2019deepmellow,patterson2022robust}. Running DQN on this environment helps identify that there may be room for algorithmic improvement.

Certain environments may start as challenge problems---ones that we do not know how to solve well---and eventually become benchmark environments. Two examples of such benchmarks are the Atari suite \citep{bellemare2013arcade, machado2018revisiting} and Mujoco environments \citep{todorov2012mujoco}. A challenge problem plays a different role than a benchmark environment. It highlights gaps in all of our algorithms; trying to fill these gaps can drive algorithm development. If an environment is truly a challenge problem, then experiments are more \emph{exploratory} or \emph{demonstrative}. It can be sufficient to show that you can obtain good performance---demonstrate something is possible---without even comparing to any other algorithms or only including basic baselines. The results suggest that you can do something that was not possible before.\footnote{This is not how Atari is now used in the reinforcement learning literature, but the first experiments including the original DQN paper \citep{mnih2013playing} were very much of this form.}

As our algorithms improve on these challenge problems, they become benchmark environments. At some point, however, they can be in a confusing interim stage. We as yet do not have an understanding of how to obtain good performance, but algorithms have gone from being abysmal to merely mediocre.

%It is typically at this stage that there becomes a single-minded push that all new work should show clear improvements in these environments, with comparisons to the now growing list of algorithms that perform okay in these environments. These environments are nontrivial in comparison to the well-known simpler benchmark problems, which we use to debug our algorithms. But they are not so hard that it is worthwhile to simply demonstrate progress; you must compare to the other algorithms and show you are better. There is a tendency to dismiss any experiments in simpler benchmark environments, since those environments are ``too easy''. This attitude misses the whole point of the experiment.

It is also at this stage that we start to overfit to these interim benchmark-challenge problem environments. The community tends uses these benchmark-challenge problems to gatekeep new algorithms: requiring new work to show clear improvements in these environments, with comparisons to the now growing list of algorithms that perform okay in these environments. There is a tendency to dismiss any experiments in simpler benchmark environments, since those environments are ``too easy''.  Eventually, this causes overfitting to the benchmarks to eke out ever smaller wins. It is an issue that happens in reinforcement learning, as well as on benchmark datasets in machine learning.\footnote{And machine learning is, of course, not the first place to experience these pains. Benchmarking also became a serious issue in planning and search \citep{hooker1995testing}. A particularly evocative quote is as follows. ``It would be absurd to ground structural engineering, for instance, solely on a series of competitions in which, say, entire bridges are built, each incorporating everything the designer knows about how to obtain the strongest bridge for the least cost. This would allow for only a few experiments a year, and it would be hard to extract useful knowledge from the experiments. But this is not unlike the current situation in algorithmic experimentation. Structural engineers must rely at least partly on knowledge that is obtained in controlled laboratory experiments (regarding properties of materials and so on), and it is no different with software engineers.'' \citep{hooker1995testing}}

This direction is typically not beneficial for general algorithm development. Instead, it may be better to acknowledge this transition and begin using these environments as benchmarks sooner. This means it is not key to outperform whatever (overfit) solution is considered the current state-of-the-art, but rather to take a useful baseline with well-known performance as a sanity check on your algorithm.
%For example, you might take Soft Actor-Critic as a reasonable baseline in Mujoco, because {\color{red} todo, not sure exactly what to say}.

For all the above reasons, it is important to remember that most of our experiments in benchmark environments are to identify issues with our algorithms, rather than to make bold claims about state-of-the-art performance. It is clearly useful to identify and fix issues in our algorithms using experiments. It is not as clearly useful to rank algorithms based on performance in small simulation environments. If there are stark and meaningful differences in well-known environments, then that success can and should be reported, with hopefully accompanying experiments to understand why. Inability to get stark improvements on benchmark problems, however, should not prevent pursuing an idea, nor gatekeeping others in pursuing ideas. Your experiments should highlight one setting where your new algorithm is demonstrably useful with a clear explanation of why. You can also demonstrate acceptable performance on benchmark environments, to show nothing is obviously broken.
%{\color{red} AdamC: maybe we should have a flow chart of how an idea might move from diagnostic, to benchmark environments and the looping back of algorithm development.}

\subsection{New (or modified) environments are not always better}

Designing environments is hard. It is also arguably something many of us have little expertise in. We learn a lot about developing algorithms, much less about developing environments. It is important to realize there are relatively few well-known environments that have stood the test of time, and the ones that do have been refined and fined tuned over years, if not decades. For example, Mountain Car was first proposed as the Puck on a Hill environment by Andrew Moore, in his PhD work \citep{moore1990efficient}. The environment featured continuous actions and a non-zero reward for reaching the goal. Years later, Sutton and Barto changed the dynamics representing the hill as a cosine wave, made the actions discrete, and the reward -1 per step. This later version became the standard for over 20 years, before AIGym \citep{brockman2016openai} introduced an aggressive episode cutoff of 200 steps to improve the performance of neural network learners.

Take pause when deciding to invent a new environment and consider the costs. First you need to justify clearly why we need yet another environment. The biggest concern is that you may very well invent an environment that is invalid or needs further improvement. But practically speaking, you are making more work for yourself. If you use your new environment to highlight the merits of your new algorithm, then you must retune baseline algorithms to avoid reporting results with untuned baselines, as we have already discussed in Section \ref{sec_untuned}. This can be a lot of work and is error prone. Exercise similar caution when modifying existing environments.
%A benchmark environment can be changed significantly from modifications to the reward, or the cutoffs. For example, Mountain Car has been modified to {\color{red} TODO}.
These changes need to be justified, particularly as it means that older benchmark performance no longer applies, losing one of the key reasons to use a benchmark environment in the first place. Sometimes the easy road and the more scholarly choice is to use an existing environment.

These changes, or new environments, also might cause us to accidentally design environments in support of our algorithms. For example, we might run DQN on discrete-action Pendulum swing up, and find under standard configurations it performs surprisingly poorly. We then might try a few changes, to maintain most of the essence of the environment but make learning more feasible. One such change could be introducing episode cutoffs and random start states to facilitate exploration. Now we have a new environment that has potentially been tuned to be easier for algorithms like DQN. If we test a policy gradient algorithm, or a completely new approach, then we may have inadvertently favored DQN. {\bf Co-evolving our environments and algorithms in this way is always dangerous.}
Note, this is not the same as designing new diagnostic MDPs. A diagnostic environment allows us to highlight the key issue being tackled.
%In some cases, this issue may not have been noted before, and there are no existing diagnostic MDPs for that issue. For example, to understand the bias in actor-critic algorithms that had not been noted before, a simple state-matching MDP with $s \in [0,1]$ and reward $r(s,a) = -(a-s)^2$ was introduced \citep[Figure 1]{thomas2014bias}.
It is common for papers to introduce new diagnostic environments---that may never be used again---to make a clear conceptual point.

One route for environment creation is to consider \emph{environment collections}. For example, we may collect several classic environments like Mountain Car, Cartpole and Acrobot into a Classic Control set composed of environments with non-imaged based, low-dimensional observations. Or we may collect several Atari games that seem to be more difficult in terms of exploration into an Atari Exploration set.\footnote{Even when just aggregating environments, we need to be careful that we have designed the collection to do what we think it does. For example, \citet{taiga2019benchmarking} showed how prior work focusing on so called hard exploration games in Atari caused researchers to miss that simple $\epsilon$-greedy based algorithms were actually better than count-based algorithms across the whole Atari suite. Environment design is hard.} We can then report performance in aggregate, across the set, as one macro-environment, without explicitly considering performance on each environment in the set.

There are several benefits to this approach. This design tests the performance of the algorithm on a set of related environments with a particular property, making conclusions less specialized to one specific environment.
%Mainly, it shows that some emergent phenomenon from an algorithm (e.g., decorrelated features) arises across several environments.
Of course, we still need to understand \emph{why} our algorithm behaves the way it does. Using more environments does not allow us to conclude our algorithm will perform better, but it does at least show that the emergent phenomena arises in more than one environment. Additionally, we obtain this ability to test across environments with minimal increases in compute over testing in one environment. That is because we can do fewer runs within each environment, to still get a large number of runs in the macro-environment (collection). A related strategy, to test generalization, was proposed earlier by having systematic parameters that could be varied for an environment \citep{whiteson2009generalized}, such as gravity in Mountain Car.

\begin{Remark}{}{aggregating_env}
Considering macro-environments has some benefits, but there are also challenges in aggregating performance across environments (e.g., dealing with different reward scales). %Be careful when looking at the performance in individual environments in the suite, which needs to be approached with caution if we only performed a small number of runs per environment.
We discuss this topic in more depth in Appendix \ref{sec_aggregating}.
\end{Remark}

%ADAMC: cut for space
%Let us share our own personal experience attempting to procedurally generate random MDPs, for an empirical study of off-policy algorithms \citep{white2016investigating}. A natural goal is to compare methods across a set of MDPs, to avoid specializing conclusions to one particular MDP. To do so, we generated random tabular continuing MDPs with different reward functions and connectivity. After using these MDPs, we more carefully examined the true values for each one in the set and found a disconcerting conclusion: the values in most states were very similar. In retrospect, this made sense: for a small set of well-connected states, the agent sees similar returns from most states. The unfortunate conclusion, however, is that these MDPs were not effective for testing the off-policy algorithms: bias unit could do very well without any function approximation.

%ADAMC: cut for space
%Environment design, however, is important. Things that are worth doing are often hard. The purpose of this section is to highlight that developing environments is no small undertaking. It also motivates why many chose to use games or simulators from industry, because the environment has been specified by others and is likely non-trivial. Restricting ourselves to such environments, however, is limiting. Developing environments is useful to the community; potentially this is an open area where some researchers could specialize and have big impact.
\newpage
\begin{summary}[Key insights: environment selection]{}

\begin{enumerate}[leftmargin=7pt]
\item Small diagnostic environments can be used to highlight the properties of a algorithm---called issue-oriented research. These diagnostic environments can provide conceptual clarity as well as a critical sanity check.
%\item Most of our environments are not challenge problems anymore; they are synthetic environments that we use to understand our methods. It is arbitrary to believe that Lunar Lander is not interesting, but Atari is. Such a gatekeeping mindset risks missing the main purpose of experiments.
\item Designing environments is hard; do not assume a published environment is correct or sensible.
\item %Ramifications on agent behavior can be large from even minor changes to the environment---such as adding random starts or adding episode cutoffs.
Proceed with caution when changing an existing environment and make sure you have a good justification for not using the original standard.
\item It is not better to run on more environments than less environments. It depends on the empirical question being asked. If you find yourself thinking you should run on one more environment, then make sure you ask yourself why and select the next environment deliberately rather than to simply increase the number of environments.
\item It is unlikely your algorithm will perform best on all environments. But, it should have the advantages you claim it has.
%Consider making a macro-environment (group of environments) for which your algorithm should be better, and another macro-environment where it may be worse.
Verify if your hypothesis is correct, and report both this good and bad performance.
%ADAMC: I removed the following bullet because we never talked about it above---following the rule never introduce new ideas in a conclusion
%\item Grouping environments adds structure to an otherwise mis-mash of results showing good and bad performance. However, do not post-hoc group environments based on performance. Groupings should be performed beforehand based on some criteria, to test a hypothesis. Changing the hypothesis after the fact is called \emph{harking} and results in instant shame, especially from psychologists.
\item It is not acceptable to run an incomplete experiment due to insufficient resources or compute. The choice of environment non-trivially impacts the cost of an experiment. Design and run a meaningful experiment in an environment that is feasible for the resources you have available.
\end{enumerate}

\end{summary}
%ADAMC: Removed summary as i think it was not needed
%In the end, we generalize our conclusions from the small set of experiments we conducted.
%We can only make limited claims about the generality of the method, even if we run on many environments. In fact, we can only make more measured claims about the outcomes we observe, under the particular conditions in which we test the methods. This includes the particular set of environments, and the hyperparameter selection strategy, and how many steps of interaction were used, and so on. If designed well, however, such measured claims can be highly meaningful: the key outcome is improved understanding of the properties of an algorithm. Insight-driven outcomes should generalize. These insights help direct further algorithm development and experiments. This contrasts with demonstrations where the goal is to show a proof of existence, without necessarily understanding the reasons explaining the observed performance---it is harder to generalize these outcomes. This is one of the reasons it is so critical to specify a hypothesis and carefully test it: to make our experiments more meaningful.

\biblio

\section{Case study: re-evaluating previous work}\label{sec:case-study}
In this section, we attempt to reproduce a well-known result from the literature, along the way evaluating the original design choices and suggesting alternatives. In the end, we draw very different conclusions!
Specifically, we attempt to recreate the experiments of the Soft Actor-Critic (SAC) paper \citep{haarnoja2018soft}.
SAC is an actor-critic algorithm derived from the maximum-entropy reinforcement learning framework and has been
shown to perform well on continuous control and robotic control tasks \citep{haarnoja2018soft,haarnoja2019soft}.

\begin{figure}[htpb!]
	\centering
	\centering
	\includegraphics[width=0.5\linewidth]{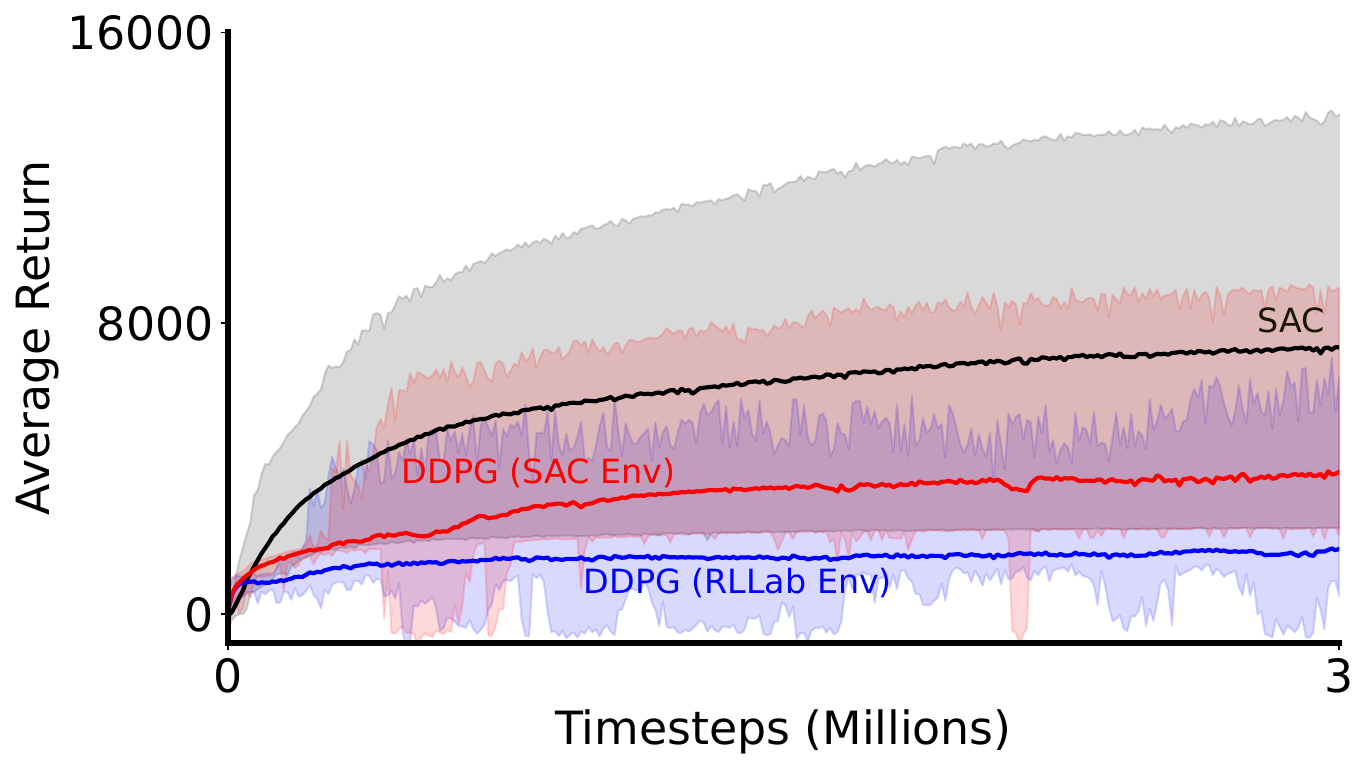}
	\caption{Our first attempt to recreate the experiments in the original SAC paper. In contrast to the original work
	which used only 5 runs, we use 30 runs. The variance in performance is higher and the average performance lower than
	reported in the original work. Solid lines denote average performance, and shaded regions denote minimum and maximum
	performance.}
	\label{fig:recreate_sac}
\end{figure}

We attempted to reproduce the comparison of SAC and Deep Deterministic Policy Gradient (DDPG) on the Half-Cheetah
environment from AIGym. For the SAC implementation, we used the original SAC codebase\footnote{The original SAC codebase can be found at
\url{https://github.com/haarnoja/sac}} (SAC-CB) with the tuned hyperparameters outlined by \citet{haarnoja2018soft}.
For the DDPG baseline, we used the RLLab codebase\footnote{The RLLab implementation can be found at
\url{https://github.com/rll/RLLab}} (RLLab-CB) with the hyperparameters set to the defaults in the RLLab codebase,
except that the batch size, replay buffer capacity, and network architectures were adjusted to match those used by SAC
\citep{duan2016benchmarking}. The SAC paper did not outline the exact hyperparameters used for DDPG, but their github
repository left some hints on the configuration of the DDPG baseline as well as the implementation used\footnote{See
\url{https://github.com/rail-berkeley/softlearning/issues/27}. It is mentioned that the original paper used the RLLab
implementation of DDPG with similar hyperparameter settings as SAC where applicable. The RLLab implementation uses OU
noise for DDPG.}. Two different environment wrappers for OpenAI Gym exist in these codebases, one in RLLab-CB and one in
SAC-CB. Which wrapper \cite{haarnoja2018soft} used for their experiments with DDPG is unclear, so we use both for
DDPG. For SAC, we assume the wrapper in SAC-CB was used.
 
We identified a possible bug in the environment wrapper for RLLab-CB, causing this wrapper to handle episode cutoffs improperly. 
Some algorithms in RLLab-CB can bootstrap on episode cutoffs. Such a practice
considers the final state due to the episode cutoff as a terminal state due to reaching a goal
---considering the cutoff state to have a value of 0 in the TD-error.  Even if this functionality is explicitly turned off,
this incorrect bootstrapping can still occur\footnote{This happens if the wrapped AIGym environment uses a number of steps per
episode less than or equal to the
% default
number of steps per episode in the RLLab-CB environment wrapper.}. We cannot know for sure, but DDPG may have
had a significant disadvantage in the original experiments. In Figure~\ref{fig:recreate_sac}, the line labelled
\textit{DDPG (RLLab Env)} does suffer from this innocuous bug while the line labelled \textit{DDPG (SAC Env)} does not.

Figure~\ref{fig:recreate_sac} shows the mean learning curves with shaded regions as minimum and maximum performance---as done in the original work.
% Before moving to our next attempt to reproduce the result,
The mean performance of SAC over 30 runs is lower than that reported in the original
paper, and the error bars here are larger than those reported in the original paper. Also, there is a performance difference between the two DDPG results, possibly due to the bug. Next, we examined the code-base
more carefully, to understand why SAC is underperforming.

In our previously described experiment, we attempted to reproduce the results of \cite{haarnoja2018soft} using the
experimental procedures described in the paper alone.
% We attempted to reproduce the results reported by \cite{haarnoja2018soft} using the algorithms and experimental
% procedures described in the paper,
%We encountered a number of difficulties when attempting to recreate the results reported in the original work
%\citep{haarnoja2018soft}.
Yet, several implementation choices in the code-base were not reported in the paper. First, the default
implementation of some policies in SAC-CB use regularization (e.g., Gaussian policies). Second, many code examples in
SAC-CB normalize actions to stay within the environmental action bounds. Finally, several code examples in SAC-CB use an
initial random exploration phase --- actions are sampled from a uniform distribution over actions for the first 10,000
steps. Policy regularization, action normalization, and random initial exploration are not reported in the paper.
We expected initial random exploration phase was the most likely culprit for the disparity in performance reported by \cite{haarnoja2018soft}, simply because
regularization and action normalization do not affect the squashed Gaussian policy implementation in SAC-CB,

\begin{figure}[htb]
	\centering
	\includegraphics[width=0.5\linewidth]{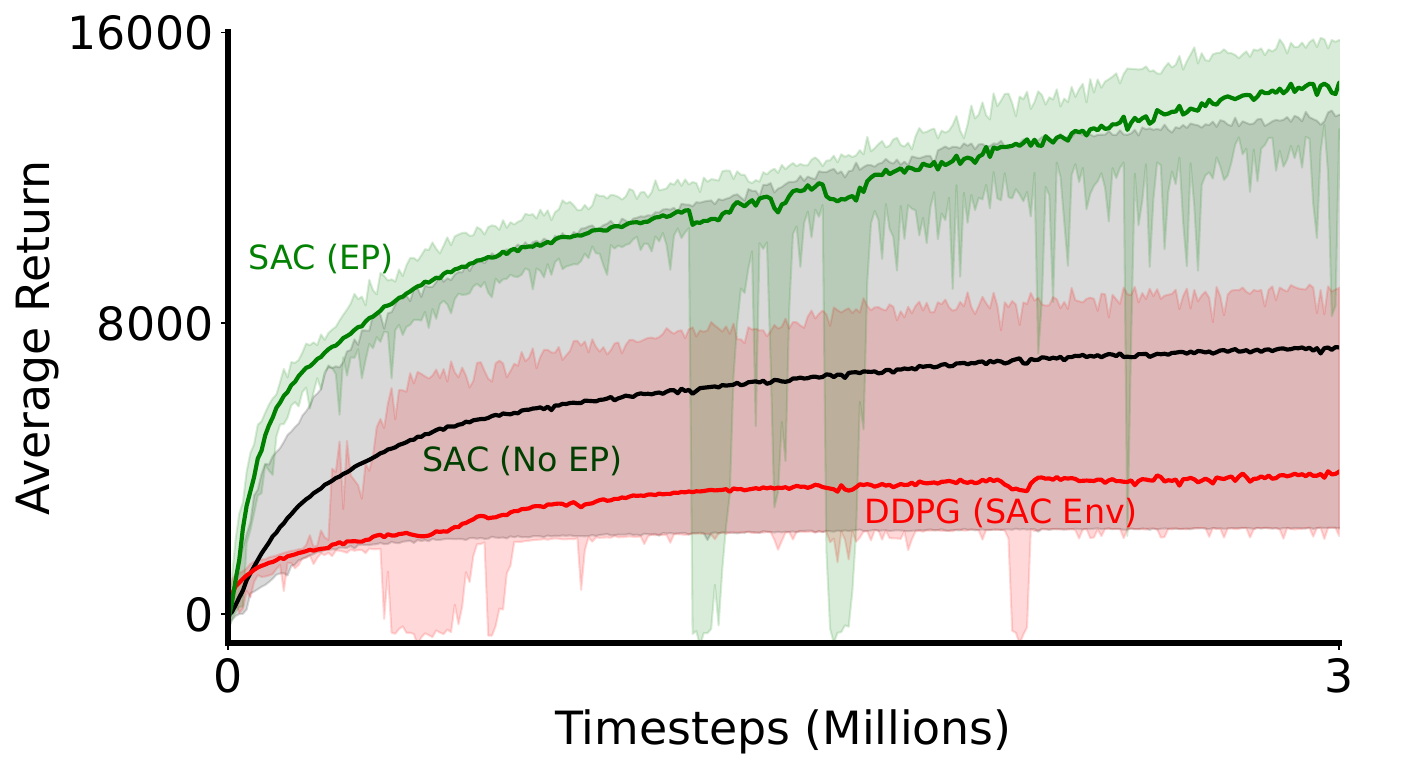}
	\caption{Our second attempt to recreate the experiments in the original SAC paper. We again use 30 runs.
		The mean performance attained by SAC (EP) closely matches that reported in the
		original work, but the shaded regions are not as tight. This could be due to the fact that we used more
		seeds than the original work to evaluate the performance of SAC.
	}
	\label{fig:recreate_sac_uniform}
\end{figure}

We re-ran the previous experiment using an initial exploration phase of 10,000 steps. 
%After this initial phase, action selection
%once again became on-policy, with actions selected according to the policy learned by SAC. 
From here on, we refer to an
agent $\mathcal{A}$ with this initial exploration phase as $\mathcal{A}$ (\textit{EP}) and without this initial
exploration phase as $\mathcal{A}$ (\textit{No EP}).
Figure~\ref{fig:recreate_sac_uniform} shows the learning curves over 30 runs for this additional variant of the SAC
agent, SAC (EP). The mean performance of this variant closely matches that reported by \cite{haarnoja2018soft}, although the
variability in performance is still noticeably higher.  This could be due to the fact that we used 30 seeds while the
original work used only 5.

\begin{figure}[htb]
	\centering
	\includegraphics[width=0.5\linewidth]{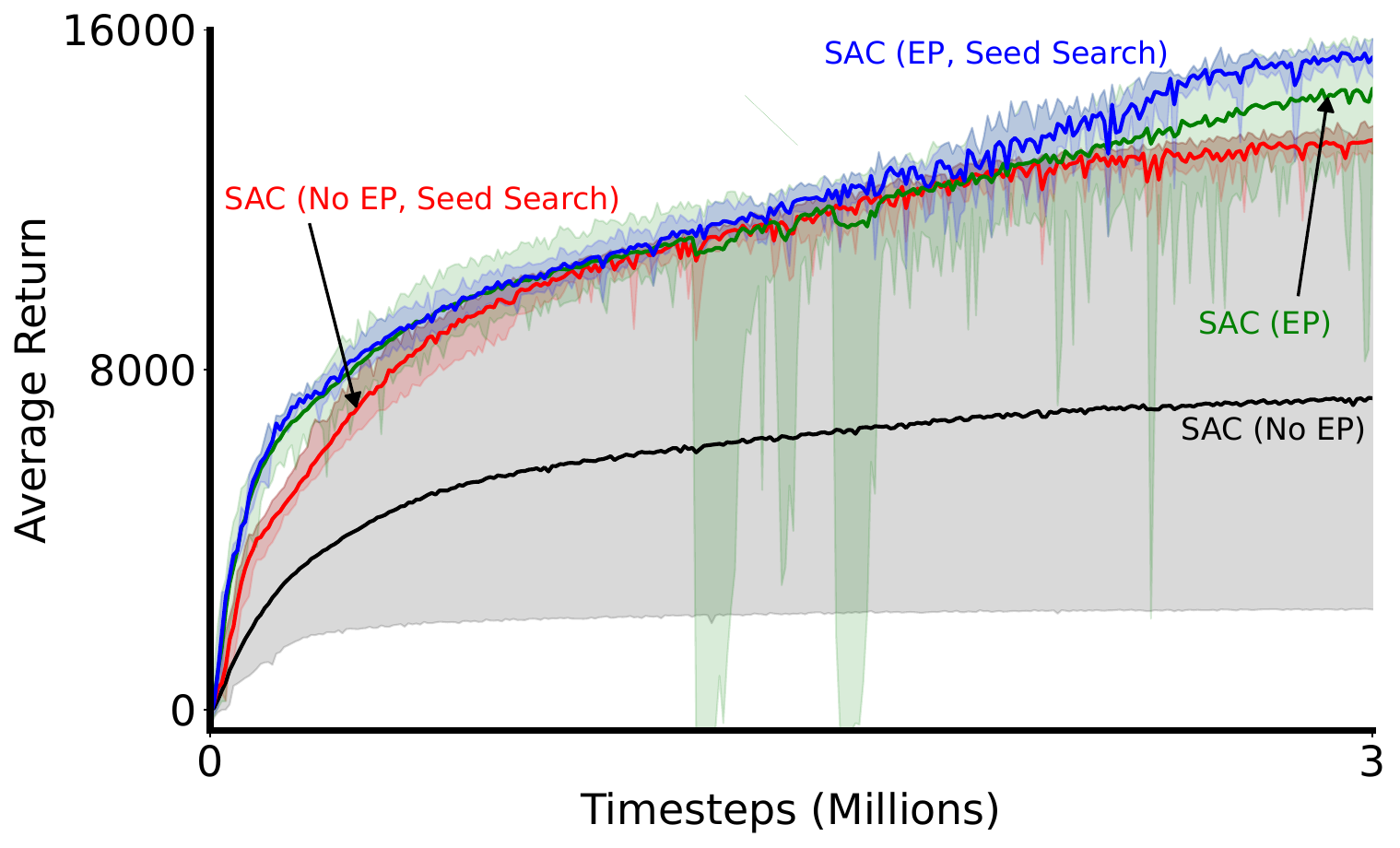}
	\caption{When searching seeds, we were able to achieve average performance and shaded regions much closer to that reported in the original SAC work. Solid lines denote average performance, and shaded
	regions denote minimum and maximum performance.}
	\label{fig:sac_only_seed_search}
\end{figure}

Finally, previous work highlighted seed optimization in the RLLab code-base, meaning that results are reported by sweeping over
seeds and reporting performance only for the best seeds \citep{islam2017reproducibility}. This seed optimization code is
compatible with SAC-CB as well. As a final attempt to reproduce the results of \cite{haarnoja2018soft}, we
used seed optimization in the hyperparameter tuning process. {\bf As a note, this is bad practice; we only conduct seed search in the
name of reproduction.} In Figure~\ref{fig:sac_only_seed_search}, where we chose the best 5 seeds of 30 for each
agent, the results more closely match those from the original paper. In particular, SAC (EP)
with seed optimization most closely matches the results reported by \cite{haarnoja2018soft}.

We now turn to running an experiment that more closely resembles the principles laid out in this document, especially
with respect to reporting performance of {\em tuned baselines}.  Although we do not know how \cite{haarnoja2018soft} tuned the
baseline algorithms in their experiments, we found that the performance of DDPG on Half Cheetah was under-reported in
this work.
%Likely, this is due to the use of default hyperparameters and lack of hyperparameter tuning for the DDPG baseline,
%which is often the case for baselines.
In the experiments here, we use the tuned hyperparameters for DDPG as
reported by SpinningUp baselines\footnote{See https://spinningup.openai.com/en/latest/spinningup/bench.html}.
Since Gaussian noise is known to outperform Ornstein-Uhlenbeck (OU) noise in some cases \citep{fujimoto2018addressing}\footnote{The benefits of
Gaussian noise were published in parallel with the SAC paper in 2018. It is therefore reasonable to assume that
the authors of SAC did not know the advantages of Gaussian noise.}, we use uncorrelated, unbounded Gaussian noise for
action exploration in DDPG instead of OU noise.
Furthermore, we try both SAC and DDPG with an exploration phase at the beginning of the experiment, where an
action is drawn uniformly randomly for the first 10,000 steps. Similarly to previous experiments, we use the tuned
hyperparameters reported by \cite{haarnoja2018soft} for SAC.
% \begin{figure}[htpb!]
%     \centering
%     \includegraphics[width=0.6\linewidth]{figures/sac/seed_search}
%     \caption{When searching seeds, we were able to achieve average performance much closer to the average reported
%     performance in the original SAC work. Solid lines denote average performance, and shaded regions denote minimum
%     and maximum performance.}
%     \label{fig:sac_lucky_seeds}
% \end{figure}

\begin{figure}[htpb]
	\centering
	\includegraphics[width=0.5\linewidth]{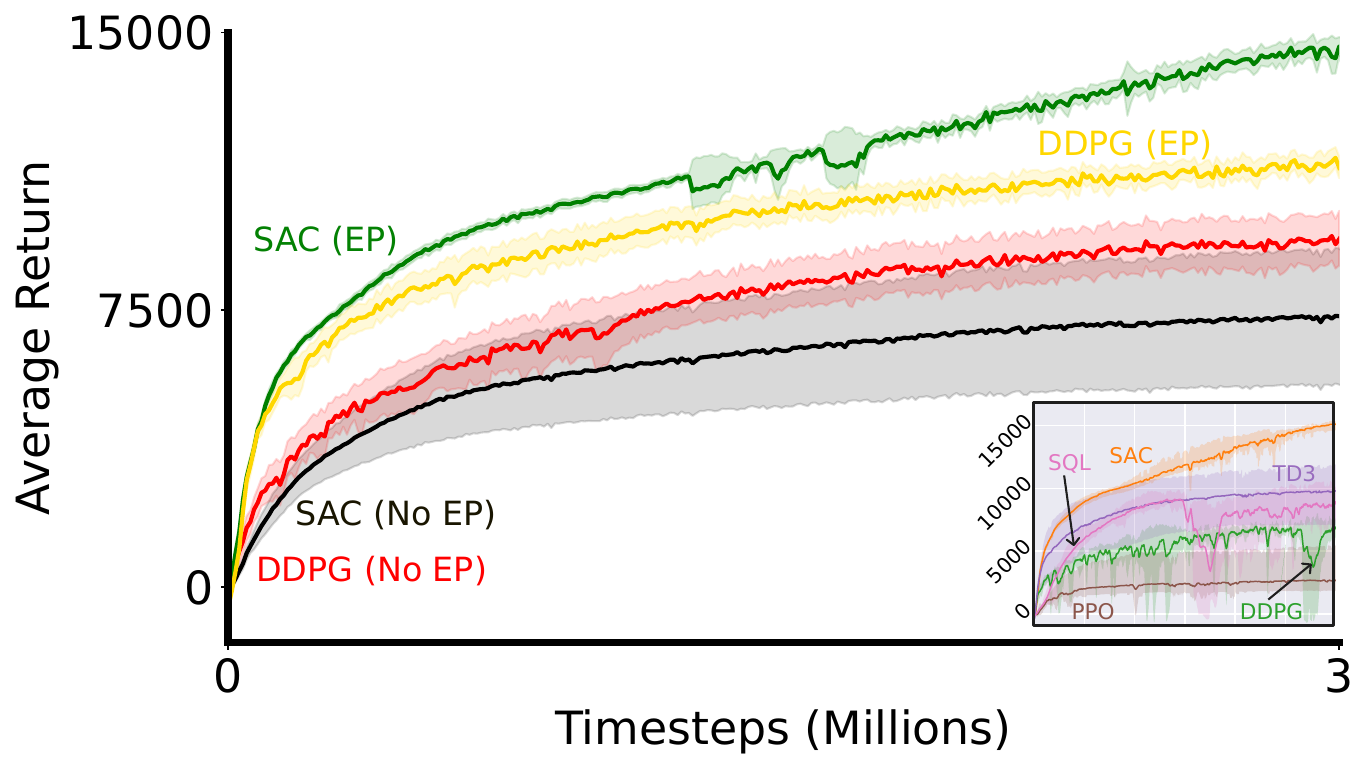}
	\caption{Our attempt to run an experiment that closely resembles the principles laid out in this document,
	particularly with respect to tuning baselines. For reference, the original experiments \cite{haarnoja2018soft}
	conducted with SAC on Half Cheetah have been inset. DDPG has been tuned here for Half Cheetah. See text for
	details.}%
	\label{fig:better_sac_exp}
\end{figure}

Figure~\ref{fig:better_sac_exp} shows the mean learning curves with 95\%
bootstrap confidence intervals for this tuned version of DDPG and SAC over 30 runs.
By tuning the DDPG baseline, we achieved significant improvements in performance over what was reported in the original
SAC work.
It seems \texttt{No EP} DDPG is competitive with and likely better than \texttt{No EP} SAC on Half Cheetah. Furthermore, we see that simply using an
initial exploration phase can significantly improve the performance of both agents on Half Cheetah.

\section{Common errors in reinforcement learning experiments}
\label{errors}

We conclude this paper by outlining a list of common errors in reinforcement learning experiments. Up to now, this document has focused largely on best practices. Naturally, it is jarring to focus too much on all the wrong choices that could be made. But both positive and negative examples of empirical design are needed to become bester empiricists. Nothing in life is black and white; what we identify below are things that generally should be avoided, but sometimes they might be appropriate. Finally, this is meant to be an ever evolving list that we will update continuously in an accompanying online blog post.

\paragraph{Averaging over 3 or 5 runs:}
sometimes it is OK to use as few as three runs. For example, if the agent and environment are both deterministic. In the vast majority of cases this is a risky practice. Even if we compute the variance across runs we cannot be sure we are not simply under-estimating the variance due to (good) luck.

\paragraph{Reusing code from another source (including hyperparameters) as is:}
qe are often tempted to simply download some code and run it to get a baseline for our plots or, even worse, simply reuse the data generated by the previous authors.
People often defend this practice claiming it is \emph{fair} and cheap, however, this only makes sense if
  (1) the implementation is trusted,
  (2) you know the performance is representative of the baseline algorithm, and
  (3) the comparisons that you are making are either a simple alteration of the prior code or you are comparing complete systems.

In the first two scenarios, if the code was written by the authors of the baseline algorithm, then you can typically trust that system-level design decisions will match their paper---though not always, see \citet{engstrom2019implementation}.
The third condition is about isolating your changes to ensure differences in performance are due only to your changes (or noise), and not due to unrelated system-level changes.
Naturally, if your scientific question is \emph{about} system-level changes, then that third condition does not apply!

\paragraph{Untuned agents in ablations:} we may take an existing agent and ablate components. The agent was potentially tuned \emph{with} those components, and there is no reason to believe the same hyperparameters will be reasonable for the agent with components removed or added.
%Ignores the covariance in performance between different hypers. Perhaps show interactions between hypers using a modelling approach instead?

\paragraph{Not controlling seeds:}
in simulation, we get to control all sources of randomness---we should take advantage of that!
When we compare two experimental conditions (for instance comparing two algorithms), we should try to minimize the number of differences between the conditions: make sure random weight initializations are the same, make sure initial state in the environment is the same, make sure the same indices are sampled for the replay buffer, etc.
These sources of randomness can amount to a huge amount of variance, but if both experimental conditions share some of the same variations then fancy analysis techniques can leverage this joint variance allowing for statistically significant claims with fewer runs.

\paragraph{Discarding or replacing runs:} sometimes the agent's parameters will diverge in one or more runs. Do not just remove this run or run it again with a different random seed! Why did it diverge? Were the hyperparameters too aggressive and this is bad performance? Maybe the agent learned so well it broke the environment. In the case you known failure is due to bad performance, then you could report $-\max$ performance. This should be done with care, especially if you are reporting normalized performance.
 
\paragraph{Cutting off episodes early:} terminating episodes early
%(say after a 100 or 200 steps, when the optimal policy $\ll$ 100 steps)
can have a large impact on exploration (making the problem easier) and likely the distribution of performance. If cutoffs are used, then they should be set to be relatively large. For example, in Mountain Car, an aggressive cutoff is 200 steps; a much more reasonable option is at least 10,000 steps.

\paragraph{Treating episode cutoffs as termination:}
setting $\gamma=0$ and making an update when the episode is cutoff is incorrect. Consider a cost to goal problem with reward of -1 per step. Termination is good and states close to the terminal state have higher (less negative values). Updating states where episodes are cutoff---which may be nowhere near the goal state---incorrectly increases the value of those states. Instead, either this last transition should be discarded or the algorithm should bootstrap off the final state before the cutoff. See Section \ref{sec_steps}.

%For instance the default practices in OpenAI Gym
%
%No longer Markov because we need full history to encode (or episode time needs to be part of state, but that causes issues with generalization)
%
%Changes the problem from an approximation of a realistic setting to a totally unrealistic setting. If I have a robot balancing a pole (or realistically balancing temperature in a building), I don?t want to shut it down and reset it periodically. I want it to learn to balance forever in a continual learning setting. Clearly infinity time is impossible with finite compute, so an approximation is necessary. Cutoffs=termination is not an approximation of this.
%
%Additional motivation being exploration. This, however, is also a bit unrealistic (i.e. I don?t want to have to go reset my robot every few hours until I?m convinced it?s running properly) and we?d prefer have agents that we can just release into the wild. Perhaps this is a call to stop using cutoffs entirely
%(Note: This is already talked about in Section \ref{sec_steps}).

\paragraph{Randomizing start states:}
the start states in an episodic MDP are part of the problem definition. In many cases they are chosen for very particular regions (e.g., the bottom of the hill in Mountain Car). Exercise  caution when changing a problem specification---especially if your motivation is to make it easier for your agent.

\paragraph{HARK'ing (hypothesis after results known):}
it is best to specify the setup of your experiment ahead of time and make a hypothesis. It is important to specify the expected outcome (e.g., my new algorithm will decrease the number of steps until the first positive reward is observed). This is the scientific method.

Do not do the following. Set up an experiment. Make no hypothesis. Run the experiment and look for a performance measure where your algorithm looks best compared to the others, continually re-running the experiment (possibly on different environments each time) until your algorithm wins.
%
%Procedure: run a hyperparam sweep, pick best results, display
%
%Alt procedure: run across several domains, pick wins, display
%
%This ties-in to our discussion on bias from procedure (i) in a previous section, can be counteracted by forming a secondary hypothesis and re-evaluating
%
%Note this is also a criticism of Scott?s work. A two-stage empirical approach is necessary to prevent bias due to HARK?ing, at least without fundamentally changing the types of claims we?re trying to make

\paragraph{Environment overfitting:}
another pitfall is to produce algorithms overly specialized to a small number of environments. For example, you might take DQN and Pong and continually add new components and hyperparameters until you improve DQN on just Pong. This is unlikely to yield generally useful algorithmic innovations and little to no insight is generated. Another example is the focus on hard exploration games in Atari as extensively investigated by \citet{taiga2019benchmarking}.
%We screen novel ideas by how well they perform on the small number of benchmarks that we have.

\paragraph{Overly complex algorithms:}
generally, we are interested in the simplest algorithms that work well, with the fewest components. The alternative---a more complex algorithm with similar performance---by construction must have components that contribute nothing. The contribution is figuring out which components matter and why.
The instinct when a algorithm is not working is to \emph{add} rather than \emph{subtract}. By focusing on \emph{why} it is not working, rather than how to fix it, you are less likely to fall into this pitfall.

\paragraph{Choosing $\gamma$ incorrectly:}

one example is choosing $\gamma = 1$ for an episodic problem with only a non-zero reward at the goal. The algorithm has no incentive to terminate: an episode of length 10 or 10,000 have the same return. For cost-to-goal problems, with -1 reward per step, it is most sensible to set $\gamma = 1$, and then we can report number of steps to goal (not discounted steps to goal). The algorithm itself might still use a $\gamma$ for learning purposes.
% MARTHAC: Too harsh, no need ot make fun of our readers
%RL 101.

\paragraph{Reporting offline performance while making conclusions about online performance:}
offline performance typically means the agent is subjected to periodic test episodes where the agent is teleported to a new state, learning is paused and we measure the cumulative reward achieved. Online performance is simply cumulative reward achieved during regular operation---during exploration and learning. Under offline performance there is no penalty for exploration during learning, whereas in online learning there is and thus the agent must trade-off exploration and exploitation. \citet{machado2018revisiting} discusses this at length.
%Again RL 101.

\paragraph{Invalid errorbars or shaded regions:}
first, whatever method you used to construct intervals, you should understand what assumptions on the underlying data are being made. Second, do not just put error bars around anything. We have already shown how our error bars can be misleading. A first warning sign to look out for is if your shaded regions are tighter than the variation in the mean across the learning curve.

\paragraph{Using random problems:}
they tend to not look much like the problems we care about (e.g., random MDPs). Use with caution.

\paragraph{Not reporting implementation details:}
one should be able to use your pseudo-code to implement your algorithm. Special tricks used to improve performance are part of the algorithm and should be reported clearly.

\paragraph{Not comparing against stupid baselines:}
can a much simpler agent or even a random policy solve your environment? Does a simpler agent outperform your agent? It does not matter if prior SOTA algorithms did not use these baselines, you should!
%Be better!

% \paragraph{Bugs}
% Never trust your code or your results. Finding bugs is good; it means one less bug.

\paragraph{Running inefficient code:}

this error might not seem pertinent to our experiments. However, if you run inefficient code, it hinders your ability to run careful systematic experiments. It inadvertently causes you to change the question you want to answer. An important first step when running experiments is to ensure you have optimized your code, so that you have more flexibility to do more exploratory experiments and feasibly run the final experiments for your paper. Always profile your code!

\paragraph{Attempting to run an experiment beyond your computational budget:} it is tempting to say it is not possible to do sufficient runs, or impossible to carefully test hyperparameters, due to lack of compute. So you conclude it is only feasible to do a smaller number of runs. \emph{However, this is not a valid claim.} Let us return to our example of animal learning researchers and rats. If a lab can only afford one rat, then it is not acceptable to run experiments using only this one rat. Instead, they simply have to do different research. The same is true for us. If you do not have the resources to run a correct experiment (say in Atari), then the alternative is not to run an incorrect one. Instead, the alternative is to run a different experiment entirely, one that is feasible and meaningful.

\paragraph{Gatekeeping with benchmark problems:}
this is more about reviewing than running good experiments, however, there is such a strong connection between the two we would be remiss not to discuss the perils of benchmarking. Experimental results should be evaluated in how they contribute understanding and insight.
%Showing that an algorithm can win a race with the state of the art is a very specific thing, and all papers need not do so.
Algorithms tuned to benchmarks often have many small tricks. It is often unclear which tricks contribute to success. Demonstration results \emph{can} be helpful---especially to inspire progress---but not all (or even most) experiments should be demonstration results. The goal of empirical reinforcement learning research is to create new knowledge, not pick winners!

\biblio

\section{Conclusion}

The goal of this document was to provide a comprehensive overview of important empirical design decisions in reinforcement learning. The style is educational, with a focus on clear examples and conceptual reasoning for making sound decisions. In the first few sections we focused on outlining  best practices for evaluating reinforcement learning algorithms. We then revisited an existing result, as a case study, to highlight different conclusions in light of these best practices. Much of the work focused on what to do, with some comments about what to avoid. For clarity, therefore, we concluded the work with a more explicit list of common errors to avoid. We hope this document will be useful to help newcomers to reinforcement learning, but also provide novel perspectives for any reinforcement learning empiricists.

{
\bibliographystyle{mybibstyle}
\bibliography{paper}
}

\clearpage

\appendix

\newcommand{\Etraining}{$\mathcal{E}_{\text{given}}$}

\section{Summary of Contributions}\label{app_contributions}

The primary goal of this work is to provide a detailed treatment of good empirical practices for reinforcement learning experiments, but along the way we have made several novel contributions.
%  of this work is to systematically catalog folk wisdom in RL experiments, and reasons for choices that might not be written elsewhere, as well as to rethink some of our empirical practices by systematically writing them down. We realized for ourselves that some of our empirical practices are insufficiently justified, and that we should consider alternatives. As such, this document is written educationally, documenting this reflection.
%But, because we had to answer questions along this journey,
The main text contains a variety of new results, beyond just conceptually reasoning to justify our design choices and recommendations.
%It can be hard to find those new results, nestled amongst this conceptual reasoning.
In this section, we summarize these contributions in two lists, organized by (a) novel proposals and (2) empirical findings.
\\\\
The novel methodological proposals include the following:
\begin{enumerate}
\item (Section \ref{var-of-perf}) Tolerance intervals with median performance can better highlight the instability of an agent within a run, and variability across runs, than our standard approach of using means and confidence intervals (see Figure \ref{fig_maze_tolerance_three} versus Figure \ref{fig_ci_dqn}).
\item (Section \ref{sec_hyperopt}) An efficient algorithm for estimating idealized performance (in Algorithm \ref{}), fixing issues with maximization bias and the typical two-stage approach uses for hyperparameters.
\item (Section \ref{sec_twoalgs}) Controlling the source of randomness to reduce variance, without introducing bias. This included (a) looking at differences in algorithm performance, to enable paired testing for algorithms and (b) separating the random seed for the agent and environment to better control sources of variability (see Example \ref{example_repeated}).
\end{enumerate}

The novel empirical findings include the following:
\begin{enumerate}
\item (Section \ref{obs-exps}) The performance distributions for our agents can be skewed and bimodal (DQN on Mountain Car in Figure \ref{fig:perf-dist-demo}, DQN on PuddleWorld in Section \ref{sec_moreruns}).
\item (Section \ref{sec_steps}) Episode cutoffs can have a large impact on algorithm performance, as demonstrated in Figure \ref{fig:cutoff_means}. This result highlights that the short cutoffs used are making environments easier for our agents.
\item (Section \ref{sec_moreruns}) A demonstration that common confidence interval approaches can fail to capture the mean, for the long-tailed performance distribution of DQN on PuddleWorld (Figure \ref{fig:skew_confidence_intervals}).
\item (Section \ref{sec_hypersens}) Momentum likely does not resolve issues with divergence in TD, as a TD with momentum still diverges on Baird's counterexample (see Example \ref{example_bairds}).
\item (Section \ref{sec_maxbias}) If we run 1000 experiments of DQN on the Mountain Car environment and sweep stepsizes, target network refresh rates, and replay buffer sizes for every experiment, then approximately 96\% of these experiments will over report the average performance for the best hyperparameter configuration among those tested.
\item (Section \ref{sec_untuned}) DQN with default hyperparameters can fail on even simple environments; with a small amount of tuning, it goes from failure to very good performance on Lunar Lander (see Example \ref{example_untuned}). This demonstrated the issue with using untuned baselines and making conclusions about the utility of a new approach.
\item (Section \ref{sec:case-study}) A thorough case study applying the methodologies laid out in this work. This case study highlighted issues with reproducing a previous result, issues in that previous experimental design that lead to misleading conclusions and a final result that provided a more clear empirical picture of those algorithms (Soft Actor-Critic and DDPG).
\end{enumerate}

\section{Further experimental details}\label{app_experiment-details}
\subsection{ESARSA and the Maze Gridworld}\label{app_esarsa-maze}
For this experiment, we combine the Expected SARSA algorithm (ESARSA) using an $\epsilon$-greedy policy both as the bootstrapping target and as the behavior policy.
The agent uses tile-coded features \citep{sutton2018reinforcement} mapping the $(x, y)$-coordinates within the gridworld to a large binary feature vector.
The state, action value function estimate is a linear function of the tile-coded features.

\begin{center}
\begin{tabular}{cc}
Hyperparameter & Value \\
\hline
Tiles & 4\\
Tilings & 8\\
Stepsize & 0.1\\
$\epsilon$ & 0.2\\
Experiment length & 30k steps\\
$\gamma$ & 0.99\\
\end{tabular}
\end{center}

The environment is shown in Figure~\ref{fig:ES_PW_1R}. The objective is to learn the shortest path to the goal over repeated episodes. The actions are discrete, moving the agent a fixed amount (plus noise) in the continuous two dimensional space. Actions that would move the agent outside the bounds of the world or into a wall cause no change in the state. The observation (and MDP state in this environment) is the x,y position of the agent. There is a fixed start state and goal region, and the reward is +1 for reaching the goal region, which ends the episode, and the reward is zero otherwise. The discount is $\gamma = 0.99$. Notice that, for this reward specification, the discount needs to be less than 1 to encourage the agent to reach the goal quickly: otherwise, taking 100 or 1000 steps would result in the same episodic return. For this simple environment we designed, we know that the optimal policy can get to the goal in 15 steps, meaning the optimal episodic return is $0.99^{15} = 0.86$.

\section{Computing tolerance intervals}\label{app_tolerance}

% --> Definition of tolerance interval and how to construct
Computing a tolerance interval is simple and is independent of the underlying distribution of the data.
We define an $(\alpha, \beta)$-tolerance interval as an interval that captures $\beta$ proportion of future samples\footnote{
  Note that tolerance intervals reason about the \emph{population} and not the specific samples being used to compute the interval.
  Reasoning about the whole population allows us to make inferences about future samples, instead of simply describing past experimental results.
}
(e.g. $\beta=0.9$) with a nominal error rate of $\alpha$ (e.g. $1 - \alpha = 0.95$).
Intuitively, to capture $\beta$ proportion of future samples, we might think to report upper and lower percentiles of the sample data, such that the percentiles symmetrically capture a $\beta$ proportion of samples.
In the case that $\beta = 0.9$, this would correspond to an upper percentile $u = 0.95$ and a lower percentile $l = 0.05$.
A tolerance interval takes this a step further by including an uncertainty correction---when we have received a small number of samples, how do we know the 95th percentile of the samples corresponds to the 95th percentile of the true distribution?
Tolerance intervals, then, add a slight pessimism by widening the interval based on the number of observed samples.
As we observe more samples, the uncertainty decreases and the interval approaches the naive percentile-based approach.

To compute the uncertainty corrected percentiles, we make use of the inverse CDF of the binomial distribution.
We want to ask the question: \emph{for each sample in our dataset, does this sample lie within the middle $\beta$ proportion of the distribution?}
Our success rate for the binomial distribution, then, is $\beta$ and our accepted error rate is $\alpha$.
The inverse CDF provides the number of samples $\nu$ that do not belong to the middle $\beta$ proportion of the distribution.
We distribute $\nu$ evenly across the top and bottom of the distribution, receiving indices of the sorted data $l = \tfrac{\nu}{2}$ and $u = n - \tfrac{\nu}{2}$.
Note that when $\nu$ is odd, these indices will no longer be integers.
A common practice is to interpolate evenly between the adjacent indices, or to alternatively take the floor of the lower index $l$ and the ceiling of the upper index $u$.
The interpolation approach generally provides more accurate bounds for smaller sample sizes.

\section{More about hyperparameter selection}

In this section, we provide a more in-depth discussion on hyperparameter selection. It is a topic that could fill an entire textbook, and so we opted to keep only the most basic information in the main body of this document. Here, we dive a bit deeper, to highlight a few other more advanced points.

\subsection{Understanding hyperparameter sensitivity for multiple hyperparameters}\label{app_adv_hyper}
Characterizing hyperparameter sensitivity with only one hyperparameter is relatively straightforward; it becomes more complex with multiple hyperparameters. The issues are that (1) there is a potentially combinatorial explosion, (2) there are likely interactions between hyperparameters and (3) visualization becomes more difficult.
As yet, there is no consensus strategy for understanding the performance of a partially-specified algorithm with multiple unknown hyperparameters, but we discuss a few here.

\begin{figure}
  \centering
    \includegraphics[width=0.5\linewidth]{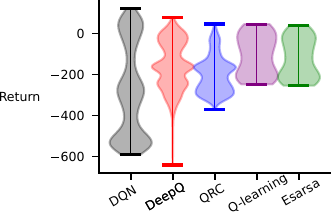}
  \caption{Visualizing the hyperparameter sensitivity of several algorithms on Lunar Landar. }\label{fig_violin_contrast}
\end{figure}

A basic strategy---that only shows variability across hyperparameters rather than interactions between them---is to use violin plots. We visualize this in Figure \ref{fig_violin_contrast}, for multiple algorithms. The idea is to select a set of hyperparameter settings, compute performance for each setting, and report the distribution of performance over all settings. This allows us to compare algorithms with different hyperparameters, and still ensure they get the same number of hyperparameter settings.

The one important nuance here is that, unlike sensitivity plots, selecting a wider range of hyperparameters can be misleading. Visually, violin plots encourage us to assess variability across hyperparameters. If we set the range to be too wide for one algorithm (say our competitor), then it may look sensitive simply because I selected an unreasonable range. Fairly selecting ranges for the hyperparameters should be easier than selecting the hyperparameters themselves, though, and violin plots are a useful tool when assessing performance with multiple unknown hyperparameters.

%\begin{figure}
%  \centering
%  \begin{subfigure}{.46\textwidth}
%    \includegraphics[width=0.8\linewidth]{figures/sensitivity/DQN_Acrobot-alpha-optimized}
%    \caption{\label{fig:optimized_sensitivity}
%      Stepsize sensitivity, with other hyperparameters optimized.
%    }
%  \end{subfigure}
%  \hspace{0.2cm}
%  \begin{subfigure}{.46\textwidth}
%    \includegraphics[width=0.8\linewidth]{figures/sensitivity/LunarLander-dist}
%    \caption{\label{fig:hyper_dist}
%      Distribution over performance (violin plot) across all hyperparameters.
%    }
%  \end{subfigure}
%  \caption{Visualizing the hyperparameter sensitivity of DQN on Acrobot. }\label{fig_violin_contrast}
%\end{figure}

To better understand relationships between hyperparameters, one plausible strategy is to collect a data set where hyperparameters are the independent variables and performance values are the dependent variables.
We can then fit a model to this dataset to describe relationships between hyperparameters as well as their relationship to performance.
Using such a strategy and a linear regression model, we could find---for instance---linear correlations between multiple hyperparameters.
This strategy is related to the AutoML and Bayesian optimization communities, though with the caveat that our interest is in the model itself while these communities generally use the model as a means to perform optimization \citep{he2021automl}.

We could additionally adopt a classic randomized-control trial approach to understanding the impact of one hyperparameter while others are left as unknowns.
With this strategy, we would test the hyperparameter of interest at multiple predefined levels (a hyperparameter sweep) while all other hyperparameters are treated as nuisance variables and, in an RCT experiment, nuisance variables are randomized over.
This strategy is highly related to a recently proposed approach \citep{jordan2020evaluating}, with the slight difference that \citet{jordan2020evaluating} treats \emph{all} hyperparameters as nuisance variables while we allow one to be controlled.\footnote{
  In fact, RCT experiments can be easily extended to allow multiple hyperparameters to be controlled.
  Such an experiment design is often called \emph{response surface methodology} or \emph{factorial design}.
}

The design methodologies discussed in this section are constantly changing as the field matures.
Currently, we discuss treating hyperparameters as unknowns while performing scientific analysis and using statistical reasoning to understand their impact.
However, as better hyperparameter optimization methodologies are developed and widely adopted, it will no longer be true that hyperparameters are strictly unknowns.
Consider the case of supervised learning. Hyperparameters may be initially unknown but a procedure for selecting them \emph{is} known: cross-validation.
As such, it is generally far more accurate and less error-prone to use cross-validation to set hyperparameter values as part of the scientific analysis instead of treating them as nuisance.
Lacking such a well-adopted strategy in reinforcement learning, in the most general case we should continue to consider hyperparameters as unknowns.

It is important also to note that tuning over hyperparameters is a luxury we have in our experiments, rather than a general-purpose algorithm to set hyperparameters.
We are not advocating here for the use of hyperparameter sweeps to select hyperparameters, as this is rarely a viable option for real-world problems, unless the environment is a simulator.\footnote{There are many groups pursuing algorithm development in reinforcement learning to solve simulation environments, like games or hard search problems. Our goal is to help empirical design more generally in reinforcement learning, not just for this more restricted problem setting, and so we do not consider specific approaches that can exploit simulators.} Instead, the goal of hyperparameter sweeps in our experiment is to get insights into hyperparameter sensitivity, and to identify algorithms that are generally performant and relatively insensitive to their hyperparameters and so more suitable for deployment.

\subsection{Developing algorithms and avoiding misleading yourself}\label{app_dev_alg}
Recall we had two possible goals in conducting an experiment: (1) understanding performance with respect to an algorithms hyperparameters, and (2) optimizing the hyperparameters for a specific problem setting. In this section we discuss something sort-of in between: selecting hyperparameters when you are developing a new algorithm. Similar to the last section, your goal here will be to understand the impact of your new hyperparameters.
The primary difference, however, is that you will likely want to iterate to improve your algorithm.
Further, you might be building on other algorithms that already have hyperparameters, and your modification to the algorithm might interact with those existing hyperparameters.

\begin{myexample}[Your new algorithm]
  In some preliminary experiments, you realize that DQN is quite sensitive to its target network update frequency. These exploratory observations inspire you come up with a new algorithm to adapt the update frequency during learning.
  Your algorithm, however, has a new hyperparameter.
  Naturally, you want to see the behavior of your algorithm under ideal conditions, where you look at performance for a nearly optimal hyperparameter setting, found either use sweeps or other (smarter) hyperparameter optimization approaches.
\end{myexample}

This is a reasonable first step---while developing an algorithm we often want to know ``does this work \emph{at all}?''---however we cannot stop our investigation here.
First, it is likely you can improve on DQN by using its same hyperparameter settings for the environment, and tuning over this additional hyperparameter.
You may falsely conclude that your algorithm provides benefits, when in actuality the main affect was the ability to optimize over an additional scalar.
Second, this modification to DQN, and your new hyperparameter, might interact with existing hyperparameter choices in an unexpected ways.
In fact, such a result has recently been shown for GANs; many modern GAN architectures provide little or no improvement over a sensibly tuned baseline \citep{lucic2018are}.

Instead, we should immediately perform the follow-up step: examine the performance of your algorithm for multiple values for the new hyperparameter.
This is a tricky and nuanced procedure, we cover many of the common cases above in Section~\ref{sec_hypersens}.
During the algorithm development cycle, it can be expensive to perform repeated sensitivity studies.
We strongly recommend that this is done on smaller, meaningful problem settings where simulation is cheap.
Such studies are often called \emph{pilot studies} and are generally used to inform the design of more complete studies later in the process.
%We cover meaningful environment selection for scientific study later in Section~\ref{sec_environments}.

\subsection{Why can't we just use cross-validation?}\label{app_cv}

A natural question is why we cannot handle hyperparameters in the same way they are handled in supervised learning: internal cross-validation. In this section, we explain why internal cross-validation does not directly apply to reinforcement learning.
% then explain one way the ideas behind cross-validation could inspire a new---but untested and speculative---approach to selecting hyperparameters and finally conclude with a simpler strategy that is a reasonable default for now.

%\subsubsection{Internal cross-validation does not directly apply to online reinforcement learning}

Let us start with a brief refresher of cross-validation. Imagine you have a dataset and want to learn a function $f$ with any regression approach. It is typical to split up this dataset into training, validation and test, where you use the validation set to select hyperparameters and the test set to get an unbiased estimate of the accuracy of the function before finally deploying. For example, if we want to select a regularization penalty $\lambda \in \{0, 0.1, 0.5\}$, we learn the function with each regularization penalty on the training set and check which of the three produces best performance on the validation set. Let's say it is 0.1. Then we learn the function on the combined training and validation sets, with regularization penalty $0.1$, to get our final function $f$. Finally we evaluate $f$ on the testing set before we deploy it into the real world. We should not have used the testing set at all during the hyperparameter selection or training phases.

This procedure, however, is not the best choice for small datasets: you want an estimate of its generalization error, but do not want to split up your already small dataset into a training and testing and/or validation sets. Fortunately, you can use cross-validation to exactly avoid this. Consider first a setting where we do not have hyperparameters, and simply want to avoid using a test set to evaluate $f$ before deployment. The idea is to partition the dataset into $k$ folds, and train on each subset of $k-1$ folds and test on the remaining fold. This procedure generates $k$ functions $f_1, \ldots, f_k$ with corresponding estimates of error $e_1, \ldots, e_k$. The average of these $k$ errors provides a reasonable estimate of the generalization error of $f$, even though they are estimates of error for different functions. \emph{The primary role of cross-validation is to estimate the generalization error of a function}, without needing a hold-out set.

The same idea can be used to select hyperparameters, in a procedure called internal cross-validation. The reason cross-validation can be used for this purpose is that it allows us to estimate the generalization error of each function learned with different hyperparameters. The goal is to pick the function with the best generalization error. The algorithm needs to both (a) specify its own hyperparameters and (b) learn its weights with regression. We can see this as an expanded learning problem, with cross-validation used as the algorithm to identify the hyperparameters. We can use cross-validation on the given dataset to evaluate each hyperparameter setting and pick the one with lowest error. This complete algorithm $A$ inputs a training set and outputs a function $f$.

In summary, external cross-validation is used to estimate test performance before deployment and internal cross-validation is an algorithm. We might use external cross-validation to evaluate $A$, which itself uses internal cross-validation. External cross-validation gives an estimate of performance, before deploying the model $f$ learned by $A$ on the entire dataset. Note that these are completely separate, and we do not have to use external cross-validation if we use internal cross-validation. For example, to evaluate the final function, we can use the more basic dataset split approach described above rather than external cross-validation. We could split our dataset into training and test, use algorithm $A$ on the training set---namely we use internal cross-validation on the training set to pick hyperparameters---and then we evaluate $f$ on the test set. Similarly, we could use external cross-validation to evaluate our final $f$ and algorithm $A$ internally could use a basic training and validation split to pick hyperparameters.

%More simply all of the above could have been described by assuming we split our dataset into training and validation. The validation set allows us to evaluate each hyperparameter, because it allows us to evaluate the generalization error of the function learned with that hyperparameter. After identifying a good hyperparameter using this approach, one would then retrain the function on all the data before deploying. The training-validation split approach is simpler, but does not provide as good an estimate of the generalization error of the function trained on all the data, especially if we have a smaller amount of data.

This brief refresher should make it more clear why
%Now let us return to the point of this brief refresher on cross-validation: why
it is not straightforward to use cross-validation in online reinforcement learning.
This idea does not directly extend to hyperparameter selection for online reinforcement learning because we evaluate a learning algorithm rather than a learned model. We can only see performance of the learning algorithm once it is in deployment (testing). There is no separate training phase, nor training data, for the online reinforcement learning setting. Overall, there is no obvious, out-of-the-box way to directly use cross-validation for hyperparameter selection.

\newcommand{\chosenh}{\hat{h}}

\subsection{Cross-validation-like procedures for the hyperparameter optimizer}

We can, however, consider strategies for selecting hyperparameters inspired by these ideas from cross-validation. We propose one such view below. Imagine we care about how our online reinforcement learning agent performs when learning from scratch on a class of environments $\mathcal{E}$---none of the below will make any sense if we only care about performance in one environment. We only receive a subset of these environments, \Etraining; let's say we have $n$ such given environments.
% MARTHAC: The below feels a bit like a bait-and-switch, since we lead them down one path and then say: nope! I've removed my own paragraph
%The goal is to answer: how will my (fully-specified) algorithm perform on $\mathcal{E}$, given only access to \Etraining?

%If our algorithm has no hyperparameters, then we can run this fully-specified algorithm in our given environments, to get samples of performance $p_1, \ldots, p_n$, and take the average. If \Etraining\ is a random subsample, then this sample average estimate is an unbiased estimate, and with big enough $n$ a relatively good reflection of performance in expectation across $\mathcal{E}$. In fact, it is more accurate to think of \Etraining\ as a testing set, because we are evaluating our algorithm using this set rather than using it for training. No training occurs here, because our algorithm does not transfer any agents from these environments. It learns from scratch on any new environment in $\mathcal{E}$.

If we have hyperparameters to specify in our algorithm, then we can consider how to use \Etraining\ to specify these hyperparameters. We can think of \Etraining\ like a training set, but not for our algorithm, but rather the hyperparameter optimizer $H$.
Each environment is like a training point. The hyperparameter optimizer $H$ can train on these environment training points, to output a proposed set of hyperparameters $\chosenh$: $H(\mathcal{E}_{\text{given}}) = \chosenh$. These hyperparameters are then deployed, and we hope that they generalize to the larger class of environments. If we had to map to the supervised learning setting, the $\chosenh$ is like the deployed function $f$ above.

The ability to generalize well depends on the quality of the set \Etraining---just like in supervised learning---and also on the hyperparameter optimization algorithm $H$. For example, $H$ could be a simple grid search or $H$ could be a Bayesian optimization algorithm. Likely these two algorithms will identify different hyperparameters using the set \Etraining, and so will result in different generalization performance. A grid search algorithm estimates the performance of each hyperparameter choice by computing per-environment performance over multiple runs in each \Etraining, and then aggregating that performance into one number. It selects (``learns'') the hyperparameter $\chosenh$ with the best cross-environment performance number.

We can use a similar idea to (external) cross-validation to estimate how well these hyperparameters $\chosenh$ might generalize. How do we know if $\chosenh$ generalizes well to other environments? We can keep a hold-out set of environments to test generalization performance. We split \Etraining\ into training and testing environments, and only give the training environments to $H$ to produce $\chosenh$. Then we can test our fully-specified reinforcement learning algorithm---full-specified because it uses hyperparameters $\chosenh$---in the test environments to get a sense of performance.

This is where the ideas behind cross-validation help. Just using training-testing splits is ``data'' inefficient:  we ``learn'' $\chosenh$ on an even smaller training set, potentially resulting in a $\chosenh$ that has worse generalization performance. Instead, we can estimate how well $\chosenh$ might do, by looking at generalization performance of all the hyperparameters trained on subsets of the environments.

More specifically, imagine we have $n$ environments. We run $H$ on all $n$ to get $\chosenh$. Now we use (leave-one-out) cross-validation to evaluate $\chosenh$. We run $H$ on all the environments except environment 1, to get $\chosenh_1$, then on all the environments except environment 2 to get $\chosenh_2$, and so on. We get an estimate of performance for $\chosenh_1$ by running the reinforcement learning algorithm with hyperparameter setting $\chosenh_1$ on environment 1, to get $p_1$. Then we do the same for $\chosenh_2$ on environment 2.  The estimate of performance for $\chosenh$ is $\frac{1}{n} \sum_{i=1}^n p_i$. If instead we pick $k < n$ folds, say if $n = 10$ and $k = 5$, then we would be training on 8 environments and testing on 2. Note that this evaluation procedure does not actually change what hyperparameters are deployed: we are still deploying $\chosenh$. It just lets us get an estimate of the quality of these before deployment.\footnote{Note that this procedure has never been used, to the best of our knowledge, and so it is not clear that it enjoys the same nice properties as cross-validation. It might have unexpected sources of bias. We are not necessarily advocating that we use the above procedure, but rather trying to bring clarity by showing potential connections to standard algorithms in supervised learning.}

% MARTHAC: Why did I put a new subsection here?
%\subsection{Even more about treating the hyperparameter optimizer as the algorithm}

The above was only using external cross-validation to estimate generalization performance. But if the hyperparameter optimizer itself has hyperparameters---let's call them hyper-hyperparameters for lack of a better term---then we can exploit the fact that the cross-validation-like approach above estimates generalization performance. We want to select the hyper-hyperparameters such that they produce hyperparameters $\chosenh$ that generalize best. For example, grid search might have a hyperparameter that is the number of runs or number of steps of interaction in the environment. These could be chosen using an internal cross-validation approach.

In summary, if we treat \Etraining\ as the dataset for our hyperparameter optimize $H$, then we can leverage ideas from cross-validation. There is no sensible mapping where internal cross-validation itself is used to select the hyperparameters of the online reinforcement learning algorithm. Internal cross-validation can be used to set the hyper-hyperparameters of our hyperparameter optimizer. If our hyperparameter optimizer does not have any hyper-hyperprameters, then we simply run the hyperparameter optimizer on \Etraining\ and deploy the hyperparameters that are found. We can use external cross-validation to estimate the quality of these hyperparameters.
%Otherwise, if the hyperparameter optimizer has 5 different hyper-hyperparameter settings that could be used, then we use internal cross-validation to pick amongst the hyper-hyperparameters.
%then we can split into training and validation. In the simplest setting, with one training set of $n$ environments and one validation set of $m$ environments, this means we would run the hyperparameter optimizer 5 times, with each hyper-hyperparameter setting, to get five possible hyperparameters. We would then test each of these 5 on the $m$ validation environments and either (a) pick the best one or (b) use the hyper-hyperparameter setting that resulted in the best hyperparameter and rerun the hyperparameter optimizer on all $n+m$ environments to get the final $\lambda_{\text{all}}$ for deployment.

It is important to point out that the premise in this section is that we have access to a set of given environments before the algorithm is deployed in an environment in $\mathcal{E}$; this requirement may be hard to satisfy. The standard assumption in supervised learning is that the training set is representative of the testing set.
% MARTHAC: This footnote is a distraction
%\footnote{Of course, there are settings with distribution shifts, between training and test. But, there is still an assumption that a training set exists and has some useful connections to the deployment (test) scenario. We have no clear mechanism to obtain training sets yet, in reinforcement learning, nor is it even clear that we should.}
In reinforcement learning, if we are deploying an algorithm to control a physical system, we do not have a set of other physical systems where we can evaluate the agent first. A practitioner might actually try to craft this set themselves. Maybe they scour the literature for environments that resemble their environment, such as environments in Mujoco or simulators for other real-world problems. They might then evaluate their algorithm---and tune the hyperparameters---in those environments. However, such a procedure has not been explicitly proposed in the literature, nor is it standard practice. It would be interesting to explore such ideas, and understand the pitfalls, to get to a similar place that supervised learning is in with cross-validation.

\section{Aggregating environments}\label{sec_aggregating}

%\subsection{Grouping environments for new benchmarks and reduced computation}\label{sec_aggregating}

Individual environments can also be grouped, to provide a \emph{macro-environment} for which we examine performance of an algorithm. We run the algorithm on each environment in the group, but consider the aggregate performance across environments.
By thinking of this environment grouping as a macro-environment, it encourages curation of reasonable groupings to test different properties of the algorithms. If we group environments with low-dimensional inputs and have a separate group for image-based inputs, then we can test the algorithms separately on these two macro-environments to understand behavior on MDPs with these two different properties.
Grouping environments, therefore, can improve on benchmarks by making experiments more issue-oriented.

\begin{figure}
    \centering
    \includegraphics[width=0.9\linewidth]{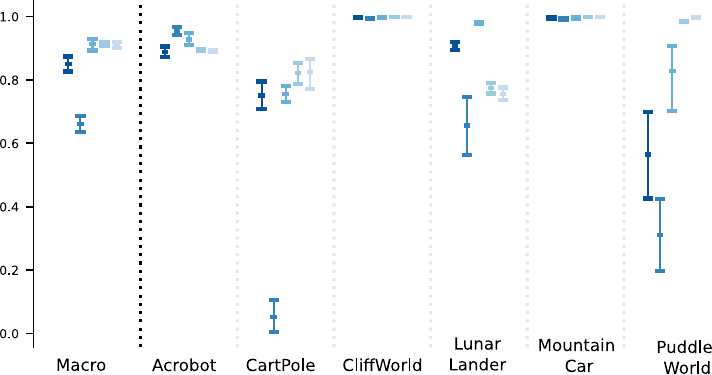}
    \caption{\label{fig:macro_env-ci}
        {\em Comparing multiple agents on multiple environments---small seed regime.} Bootstrap confidence intervals around {\bf average performance} (y-axis) across environments (labeled Macro) versus for each independent environment.
        It is difficult to assess an ordering of five algorithms when investigating environments individually, however, a clearer ordering emerges when measuring performance across environments jointly.
        Also notice the confidence intervals for the Macro environment are much more narrow.
        All intervals on individual environments are constructed with 10 samples, while the aggregate intervals for the macro-environment use 60 samples each.
    }
\end{figure}

It also has the additional benefit that we can test the algorithm in a larger set of environments, without having to do as many runs for each environment. If there are 5 environments in the macro-environment, it may be sufficient to use 10 runs for each, giving a total of 50 runs for aggregate performance in the macro-environment. Because we are not making claims about performance within each environment, fewer runs per environment are acceptable. It can still be reasonable to visualize individual runs in the environments, for qualitative insights such as those related to instability within runs.
% MARTHAC: tried to say this about individuals. Does this resolve it?
%\TODO{
%    \emph{Can} qualitative behaviors be visualized if we don't have enough runs? We certainly can't reason about behaviors which we expect generalize across runs using only 10 runs.
%    Maybe we don't mean "qualitative" here, but rather mean individual differences?
%}
Performance plots, however, should likely only be reported for the macro-environment. We visualize what performance could look like, in Figure \ref{fig:macro_env-ci} for each environment when there are too few runs, resulting in highly-overlapping errors bars, whereas aggregate performance across runs provides clear differences between the two algorithms.

% MARTHAC: Maybe this should just be talked about in the hypers section, and not repeated here. We can be more precise there.
%Grouping environments can also facilitate more reasonable hyperparameter selection. As discussed in Section \ref{sec_hypers}, it can be error-prone to tune hyperparameters per-environment in experiments, potentially providing misleading results about an algorithms performance. However, tuning hyperparameters for a macro-environment is less problematic. If there are suitable hyperparameters for all the environments in a macro-environment, then this outcome highlights that the algorithm is not extremely sensitive to its hyperparameters and that it might be tunable for related environments. It is a more realistic portrayal of algorithm performance, rather than a best-case portrayal. Of course, it can suffer from similar over-tuning issues, especially if the environments in the macro-environment are very similar; as usual when designing experiments, you need to reason about what is being shown.

Reporting such aggregate performance, however, raises the issue of how to ensure performance is normalized across environments. Such normalization could be hand-designed by the experimenter, to ensure each macro-environment has a sensible distribution. For example, the experimenter may know the optimal expected return $G^*$ and worst-case return $G^-$ in each environment. The observed returns $G_t$ could be normalized between 0 and 1 using
\begin{equation*}
\frac{G_t - G^-}{G^* - G^-}
.
\end{equation*}
If this number is 1, the return is optimal; if it is zero, the return was the worst-case.
Special care should be taken when using ratio-based scaling methods, as these can be sensitive to relative size of the scale \citep{fleming1986how,jordan2020evaluating}.

% Such an aggregation does not account for the relative difficulties of each environment, and these constants may be non-obvious to specify. An alternative approach has been to use relative performance across a set of agents evaluated \citep{andre,jordan,patterson}. {\color{blue} todo for Andy to describe this}. An issue with these approaches, though, are that numbers cannot be compared across different papers.

In addition to creating such macro-environments, there are already several environment suites that have been designed to have related but different environments. Examples include Atari \citep{mnih2013playing}, Metaworld \citep{yu2020metaworld} and MiniGrid \citep{chevalier-boisvert2018minimalistic}. Earlier work also considered minor variations of the same environment, such as Mountain Car with different perturbations to transitions, observation transformations and start-state conditions \citep{whiteson2011protecting}.

\section{Understanding our agents with multiple evaluation metrics}\label{sec_metrics}

For most of this work we have been relatively agnostic to the choice of evaluation metric. In the initial section we discussed the typical evaluation metric: the return. After seeing how to measure this evaluation metric in one run for one agent, we discussed strategies to assess an algorithm across multiple runs by aggregate this evaluation metric across runs. These aggregation strategies were generic, and apply to other evaluation metrics.

In this section, we discuss alternative evaluation metrics that could help you assess an algorithm. In some cases, the expected return is not a suitable evaluation metric. Instead, for example, you might want to know if your agent maintains a certain level of performance once reaching that level. More generally, using multiple metrics can provide a more complete picture of the properties of your algorithm. Some metrics may be behavioral, rather than performance-based. For example, it may be useful to understand state-visitation for an agent. We distinguish such \emph{behavioral metrics} from \emph{performance metrics}, and discuss options for both in the next two subsections.

\subsection{Measuring the performance of an agent}
In order to compare, rank, describe, and tune our algorithms thus far, we have described the performance of agent using the total episodic return obtained by that agent over time.
In fact, the overwhelming majority of research in reinforcement learning has focused on describing how well an RL agent solves a given problem.
However, there are many possible attributes of our agents which we can measure and many different metrics we can consider for each attribute.
A fruitful path towards understanding an agent or an algorithm is developing a multidimensional view of that agent's behaviors and internal processes.
There have been countless metrics and attributes studied in the machine learning literature---certainly far more than we could sensibly explore here---however we will describe a few important choices which can serve as a starting point.

% --> Weighted episodic return
Perhaps the most simple deviation from measuring the total episodic return of an agent is applying a weighting to each episodic return based on the number of learning steps within that episode.
Consider an environment where an agent spends ten thousand steps to complete the first episodic, such as the Mountain Car environment.
When the agent starts the second episode, it has already accumulated ten thousand experiences and has likely performed ten thousand updates to its value function.
Comparatively, an agent which solves the first episode in only a few hundred steps, it starts its second episode with far less experience and far fewer updates.
Comparing these two agents episode-by-episode means comparing agents with wildly differing amounts of experience, which may not be a meaningful comparison!
We could, instead, compare each agent at each update timestep, ignoring episode boundaries.
Our performance measure, then would need to assign a performance value at every step.
A simple proposal: for each timestep of an episode, record the total return observed for that episode.
If an agent takes fifty steps to complete an episode and yields a return of -100 at the end of the episode, then the agent would record a performance value of $[-100, -100, \ldots, -100]$ repeated fifty times---once for every step of the episode.

% --> Representation attributes
There are many interesting questions we can ask about the agent which are not directly tied to performance, such as: ``what representations are learned by our agent?''
Some attributes of interest include the capacity of the representation---what functions can learned?---the efficiency of the representation---are the duplicate features?---or the amount of interference observed by the representation for each new observation \citep{wang2022investigating}.
Although it may not be immediately clear that these additional measurements correlate with how well an agent solves a task, these can provide distinguishing information between individual agents.
Over time, and with many such measurements, we might begin to notice patterns such as: agents with high degrees of interference tend to learn quickly, but fail to adapt to small changes in the environment.
These correlations and relationships provide novel avenues for algorithm development.

\subsection{Summarizing performance over time}\label{sec_summarizing}
%It is often difficult to divorce the evaluation of an agent's performance from the temporal nature of reinforcement learning---
The performance of an agent evolves with time, as do other attributes of our agents; we want our evaluation to capture this.
On the other hand, it is likewise challenging to rank, compare, or tune algorithms when our performance metric is a stochastic process or a list of numbers.
Quite often, we need to provide a scalar summary which describes the measurements that we have taken over time.
A common choice---referred to as area under the curve---is to compute the total amount of an attribute scaled by the number of observations, the average over time.

A persistent challenge in providing a single summary scalar, however, is that nuances in the learning curve will be lost.
A time average of the learning curve is akin to fitting a horizontal over the curve, which clearly loses the fact that the curve might increase or decrease steadily with time or might have sharp drops in performance on occasion.
Time averages also ignore the fact that learning curves can often be broken into two (or more) phases, generally a rapid upward slope during the early learning process, then a progressive plateau towards the end of learning.
Depending on the research question at hand, it can often be desireable to reason about only one of these phases, for instance by computing a time average over the last 10\% of the learning curve in order to discuss where the algorithm generally plateaus.
There has been some work automatically identifying these phases, by fitting piecewise linear functions to the learning curve \citep[Chapter 3]{dabney2014adaptive}. As usual, there is no right answer, and as the experiment designer you need to make an appropriate choice for your setting.

Once you have isolated the evaluation phase, it may be reasonable to use alternatives to the average to summarize this portion of the curve.
One option is to evaluate the worst case performance during evaluation; does your agent consistently perform reasonably well or does it exhibit occasional large drops in performance?
This evaluation may be more suitable for settings where it is key to maintain reasonable performance for almost all episodes such as in medical applications where the preference is to ensure most patients get a reasonable treatment, rather than obtaining good outcomes on average across patients. The average could be maximized by having very good treatments for some patients and very poor ones for others. As another example, if the agent is learning to land a helicopter, then we expect that it should never crash the helicopter in the evaluation phase. Reporting worst-case performance in the evaluation phase, rather than average, might better reflect the desired performance for this agent.

One challenge in summarizing learning curves is that there are no agreed-upon definitions for \emph{learning speed} and \emph{stability}.
We can provide strategies to operationalize these concepts, however without concrete definitions it can be challenging to ensure our proposed summaries perfectly reflect these properties.
For instance, perhaps we define learning speed as how quickly the agent reaches a reasonable policy.
To measure this, we can define a threshold of ``reasonable'' performance---usually coming from domain knowledge---and measure how many learning steps it takes to cross this threshold.
Similarly, we might define stability as staying above that threshold of performance once it has been achieved; measured by counting how often the agent dips below this threshold in the evaluation phase.
Does the agent stay above this performance level, or does it often regress and require re-learning?

These particular measures present a challenge, however.
What if due to stochasticity our agent manages to cross our designated threshold of performance very early in learning, long before it has actually learned a sensible policy?
We might spuriously decide this agent is unstable, when in reality if we had selected a slightly later cutoff point for the evaluation phase we would have seen this agent remain stable.
To avoid this potential stochasticity, we can alter our measure to require, say, three consecutive steps above the threshold before we say the agent has crossed from the learning phase to the evaluation phase.

The choice of 3 is not optimal, nor theoretically motivated for either stability or learning speed. Instead, it is a choice motivated by the fact that we would like to be robust to stochasticity, but do not want to use too many consecutive steps as then we are measuring ``reached reasonable performance and was stable'' rather than ``reached reasonable performance''. These choices have to be made carefully, and potentially revisited. For this reason, it can be useful to provide summary performance metrics, but also provide learning curves---even showing each run---to allow for more detailed viewing for a reader that would like to dig deeper. Our job as authors is to provide insightful summaries without overloading the reader, so such detailed information may be better included in a supplement.

\subsection{Offline returns versus online returns}

We have so far exclusively considered online returns. It is common, however, to report offline returns. At each measurement step, we take the current policy and do multiple rollouts in the environment, to estimate it's current expected return. This estimate is used as the performance metric at that measurement step, plotted in the learning curve.

Reporting online returns corresponds to the online learning setting, whereas plotting offline returns corresponds to the pure exploration setting. In the online learning setting, the agent is faced with the exploration-exploitation dilemma. It is being evaluated by how much reward it receives while learning, and so it may want to take the action it currently thinks will achieve most reward (act greedily). But, it also needs to spend some time exploring, to ensure it has not settled on a suboptimal policy and so is missing out on more reward. An agent that balances these well will perform well according to online returns.

For offline returns, the agents behavior during learning is not evaluated. In that sense, it does not face the exploration-exploitation dilemma. Instead, the behavior is faced with the pure exploration question: what actions should it take to learn the target policy (near-optimal policy) on that generated data? The learning curve reflects: if the agent was able to have a pure learning phase for $t$ steps, and then deployed its fixed policy, here is how that fixed policy would perform.

Without looking carefully at a plot and its description, learning curves for these two settings can look similar. But these two settings are fundamentally different. It is important to motivate why you chose online or offline returns in evaluation.

\clearpage
\begin{summary}[Key insights: understanding agents with multiple evaluation metrics]{}
\begin{enumerate}[leftmargin=7pt]
\item Multi-dimensional analysis is key for truly understanding algorithm performance.
\item \textbf{Do} consider multiple performance metrics to understand your algorithm. You need not report all of these metrics; an exploratory phase helps guide final experiments.
\item \textbf{Do not} optimize over multiple performance metrics, simply to find which measure results in your algorithm having better performance than another. After the exploratory phase, decide what you want to measure. For final experiments reported in a paper, comparing your algorithm to others, you should consider additional/different environments from those used in the exploratory phase.
\item Offline and online returns are measuring very different things. Make the choice between these judiciously.
\item There are no clear-cut answers for defining and selecting evaluation metrics. You need to clearly justify your choices. A first criteria for the choice is whether you can convince yourself.
\item Most of our experiments should include behavioural metrics. We often test our agents in simple environments, where the goal is to understand our algorithms rather than obtain a solution in that MDP.
\end{enumerate}
\end{summary}

\biblio

\end{document}